\documentclass[10pt,twocolumn,letterpaper]{article}

\usepackage{cvpr}              

\usepackage{multirow}
\usepackage{booktabs}
\usepackage{array}
\usepackage{makecell}
\usepackage{diagbox}

\usepackage[accsupp]{axessibility}

\usepackage{graphicx}
\usepackage{amsmath}

\usepackage{verbatim}

\usepackage{bbm}

\DeclareMathOperator*{\argmin}{arg\,min}

\definecolor{cvprblue}{rgb}{0.21,0.49,0.74}
\usepackage[pagebackref,breaklinks,colorlinks,allcolors=cvprblue]{hyperref}

\def\paperID{*****} 
\def\confName{CVPR}
\def\confYear{2026}

\title{Federated Active Learning Under Extreme Non-IID and Global Class Imbalance}

\author{Chen-Chen Zong, Sheng-Jun Huang\thanks{Corresponding Author. }\\
Nanjing University of Aeronautics and Astronautics, China\\
{\tt\small \{chencz,huangsj\}@nuaa.edu.cn}
}

\begin{document}
\maketitle
\begin{abstract}

Federated active learning (FAL) seeks to reduce annotation cost under privacy constraints, yet its effectiveness degrades in realistic settings with severe global class imbalance and highly heterogeneous clients. We conduct a systematic study of query-model selection in FAL and uncover a central insight: the model that achieves more class-balanced sampling, especially for minority classes, consistently leads to better final performance. Moreover, global-model querying is beneficial only when the global distribution is highly imbalanced and client data are relatively homogeneous; otherwise, the local model is preferable.
Based on these findings, we propose FairFAL, an adaptive class-fair FAL framework. FairFAL (1) infers global imbalance and local–global divergence via lightweight prediction discrepancy, enabling adaptive selection between global and local query models; (2) performs prototype-guided pseudo-labeling using global features to promote class-aware querying; and (3) applies a two-stage uncertainty–diversity balanced sampling strategy with $k$-center refinement. Experiments on five benchmarks show that FairFAL consistently outperforms state-of-the-art approaches under challenging long-tailed and non-IID settings.
The code is available at \href{https://github.com/chenchenzong/FairFAL}{https://github.com/chenchenzong/FairFAL}.

\end{abstract}    
\section{Introduction}
\label{sec:intro}


Federated learning (FL) has emerged as a central paradigm for collaborative model training without sharing raw data \cite{mcmahan2017communication,li2020federated,karimireddy2020scaffold}, enabling privacy preservation across decentralized ecosystems such as mobile devices, hospitals, and edge--cloud infrastructures. By localizing data storage and computation, FL addresses rising privacy and regulatory constraints while leveraging rich yet siloed data sources.
Active learning (AL) aims to reduce annotation cost by selectively querying informative samples~\cite{settles2009active,huang2010active,ren2021survey}, and is especially valuable in domains where labeling is expensive, including medical imaging, autonomous driving, and industrial inspection.
Integrating these two paradigms yields federated active learning (FAL)~\cite{kim2023re,cao2023knowledge,chen2024think,ijcai2025p812}, which enables decentralized clients to collaboratively identify valuable samples without revealing raw data. This combination is particularly compelling for real-world deployments, where data are inherently distributed, privacy-sensitive, and annotation budgets remain severely constrained.

Despite its promise, FAL remains under-explored in realistic settings characterized by severe global distribution imbalance and highly heterogeneous clients---conditions that frequently arise in practice. Although recent studies~\cite{kim2023re,cao2023knowledge,ijcai2025p812} have begun to consider extreme non-independent and identically distributed (non-IID) scenarios, they often treat heterogeneity merely as a data-partitioning issue rather than a global class imbalance problem. As a result, most methods implicitly assume that the overall label distribution is relatively balanced or can be recovered through aggregation. However, real-world federated systems commonly exhibit long-tailed global distributions, where rare but critical classes appear sparsely across clients. Under such regimes, existing acquisition strategies struggle to surface minority classes, leading to systematic bias toward head categories and inefficient use of annotation budgets.


To address these limitations, we investigate a fundamental question: in FAL, two query-selection models naturally exist--the globally aggregated global model and the independently trained local model. \textbf{Which model enables sampling that better preserves class balance, particularly for rare classes, and how does such balance influence the performance of the final global model?}


Our empirical analysis yields a consistent insight: across varying levels of global imbalance and client heterogeneity, the model that achieves more class-balanced sampling reliably delivers superior final performance. Furthermore, when global class imbalance is severe but client distributions are relatively homogeneous, global-model–based sampling proves more effective; in all other regimes, the local model generally becomes the preferred choice.



Building on these insights, we introduce FairFAL, an adaptive, class-\textbf{fair FAL} strategy designed to handle extreme non-IID data and globally long-tailed class distributions. FairFAL incorporates three key components. First, we exploit the prediction discrepancies between the global and local models on each client to estimate both the severity of global class imbalance and the degree of local heterogeneity, enabling an adaptive model-selection mechanism that chooses the more suitable query selector per client. Second, motivated by the observation that the global model consistently provides stronger and more coherent representations, we compute per-class prototypes using global features and perform prototype-guided pseudo-labeling to enforce class-aware querying. Third, to reduce redundancy and promote coverage, we form class-specific candidate pools and apply $k$-center selection in a gradient-embedding space that combines global features with model predictions, ensuring that the final queries are both informative and diverse. Experiments on five benchmark datasets show that FairFAL consistently outperforms state-of-the-art FAL baselines under challenging federated settings.
\section{Related Work}
\label{sec:related_work}

\textbf{Federated learning (FL)} enables multiple clients to collaboratively train a shared model without exchanging raw data.
FedAvg~\cite{mcmahan2017communication} established the standard framework, but its performance degrades significantly under statistical heterogeneity (non-IID data).
To mitigate this, prior work has explored three main directions.
Regularization-based approaches~\cite{karimireddy2020scaffold,li2020federated,guo2023fedbr,huang2023rethinking} constrain local updates or feature representations to reduce client drift.
Aggregation-based methods~\cite{jiang2023fair,ma2023federated,rehman2023dawa,zhang2023federated} adjust client contributions based on divergence or data quality.
Personalized FL~\cite{chen2021personalized,liu2021feddg,wang2022personalizing,ye2023heterogeneous} further decouples global and client-specific components to improve adaptability across heterogeneous distributions.
Beyond the standard setting, FL has also been extended to semi-supervised~\cite{liang2022rscfed,bai2024combating}, multi-task~\cite{smith2017federated,marfoq2021federated}, continual~\cite{yoon2021federated,shenaj2023asynchronous}, and domain-adaptive~\cite{jia2024dapperfl,xiao2025unsupervised} learning, highlighting its versatility in real-world applications.


\textbf{Active learning (AL)} aims to reduce annotation cost by selectively querying the most informative samples from an unlabeled pool.
Existing methods fall into three main categories.
Uncertainty-based approaches~\cite{li2006confidence,balcan2007margin,holub2008entropy,gal2016dropout,yoo2019learning,parvaneh2022active} focus more on ambiguous samples near the decision boundary, typically measured through entropy~\cite{holub2008entropy} or margin~\cite{balcan2007margin}.
Diversity-based strategies~\cite{nguyen2004active,kutsuna2012active,urner2013plal,sener2017active,caramalau2021sequential} such as Coreset~\cite{sener2017active} focus on selecting representative points that cover the feature space.
Hybrid methods~\cite{ash2019deep,prabhu2021active,wang2023mhpl,yuan2023bi3d,zong2025rethinking} integrate both, with BADGE~\cite{ash2019deep} being a well-known example of two-stage uncertainty–diversity balanced sampling.
Recent developments extend AL to low-budget~\cite{hacohen2023select,bae2024generalized}, semi-supervised~\cite{gao2020consistency,huang2021semi}, and cross-domain~\cite{yuan2023bi3d,werner2024cross} scenarios.

\textbf{Federated active learning (FAL)} integrates AL into FL to enable label-efficient model training under privacy constraints.
Early studies~\cite{ahmed2020active,wu2022federated,ahn2024federated} directly applied centralized AL strategies using either a global or a local model, but such single-model designs degrade quickly under strong client heterogeneity.
Recent methods~\cite{kim2023re,cao2023knowledge,chen2024think,ijcai2025p812} therefore adopt hybrid designs that combine global and local signals.
For example, LoGo~\cite{kim2023re} performs local clustering followed by global uncertainty scoring, while KAFAL~\cite{cao2023knowledge} and IFAL~\cite{ijcai2025p812} exploit prediction discrepancies between global and local models to guide acquisition.


\section{Preliminary}

\textbf{Federated learning (FL)} enables a federation of clients to collaboratively train a shared model without exchanging raw data.  
Let $K$ clients hold local datasets $\{\mathcal{D}_1, \dots, \mathcal{D}_K\}$, where each sample $(x_i, y_i)$ consists of an input $x_i \in \mathcal{X}$ and a label $y_i \in \mathcal{C}=\{1,\dots,C\}$ for a $C$-class task.  
Denote the global model parameters by $\theta^g$ and the local parameters on client $k$ by $\theta_k$.  
The training objective aggregates client-level losses in a weighted manner:
\begin{equation}
    \min_{\theta^g}\; \sum_{k=1}^{K} \frac{n_k}{N}\, \mathcal{L}_k(\theta_k),
    \quad 
    n_k = |\mathcal{D}_k|, \;\; N=\sum_{k=1}^{K} n_k .
\end{equation}
In each communication round, client $k$ trains $\theta_k$ on its private data and sends the resulting model update to the server. The server aggregates all client updates, typically via FedAvg~\cite{mcmahan2017communication}, to obtain the new global parameters $\theta^g$.
This paradigm enables privacy-preserving learning across heterogeneous data silos while avoiding direct data sharing.


\begin{figure*}[t]
  \centering
   \includegraphics[width=0.97\linewidth]{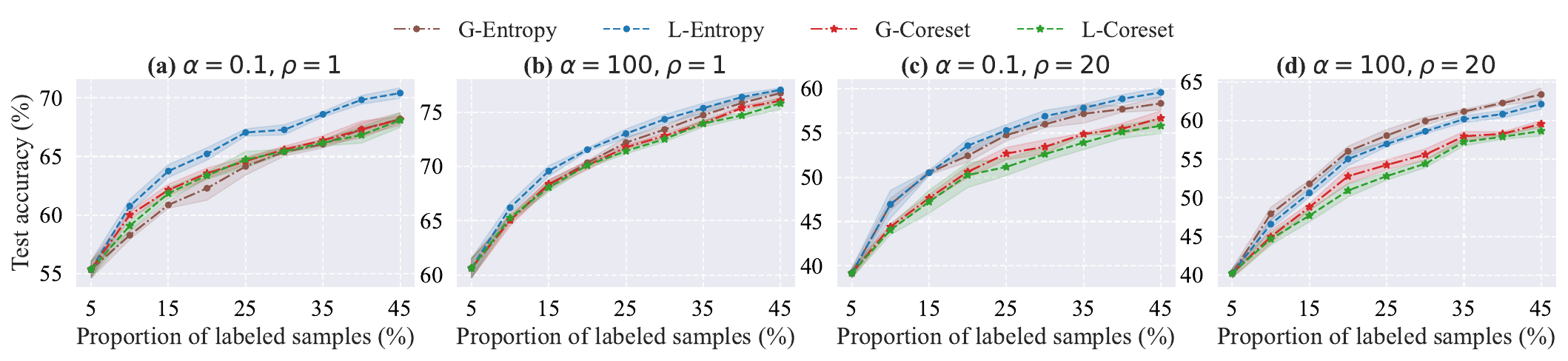}
   \caption{Comparison of global (G) and local (L) query selectors under varying $(\alpha,\rho)$ configurations on CIFAR-10. 
Test accuracy is plotted against the proportion of labeled samples for Entropy and Coreset sampling.
}
   \label{fig:c10_curve}
\vspace{-3mm}
\end{figure*}

\textbf{Active learning (AL)} aims to minimize annotation cost by iteratively querying the most informative unlabeled samples for labeling.  
Given a labeled set $\mathcal{D}_L=\{(x_i,y_i)\}$ and an unlabeled pool $\mathcal{D}_U=\{x_j\}$, a model with parameters $\theta$ is iteratively trained on $\mathcal{D}_L$ and improved through selective acquisition.  
At each AL round, the learner selects a query set $\mathcal{Q} \subset \mathcal{D}_U$ using an acquisition function $\mathcal{A}(\cdot)$ that scores the informativeness of each unlabeled instance.  
The queried samples are labeled and added to $\mathcal{D}_L$, while the unlabeled pool is updated as $\mathcal{D}_U \leftarrow \mathcal{D}_U \setminus \mathcal{Q}$.  
This iterative procedure continues until the annotation budget is exhausted, enabling label-efficient model training.


\textbf{Federated active learning (FAL)} integrates the privacy-preserving collaboration of FL with the label-efficiency of AL, enabling decentralized clients to identify globally informative samples without sharing raw data. 
Unlike conventional AL, which operates with a single centralized model, 
FAL performs a full federated training process in every AL round.
Formally, consider $K$ clients, each initialized with a labeled set $\mathcal{D}_L^{(k)}$ and an unlabeled pool $\mathcal{D}_U^{(k)}$. 
At each AL round $t$, each client trains its local model on $\mathcal{D}_L^{(k)}|_{t}$ for several communication rounds, after which the server aggregates the local updates to obtain the new global model.  
Once federated training completes, each client scores its unlabeled pool $\mathcal{D}_U^{(k)}|_{t}$ using an acquisition function $\mathcal{A}^{(k)}(\cdot)$, selects a query set $\mathcal{Q}^{(k)}|_{t} \subset \mathcal{D}_U^{(k)}|_{t}$, and acquires their labels.  
The labeled and unlabeled sets are updated via
\(\mathcal{D}_L^{(k)}|_{t+1} \leftarrow \mathcal{D}_L^{(k)}|_{t} \cup \mathcal{Q}^{(k)}|_{t}\) and
\(\mathcal{D}_U^{(k)}|_{t+1} \leftarrow \mathcal{D}_U^{(k)}|_{t} \setminus \mathcal{Q}^{(k)}|_{t}\).
This process alternates between federated training and active querying until the annotation budget is exhausted, gradually improving the global model with minimal supervision.


\section{Observation}
\label{sec:observation}

We begin by investigating how global and local models perform as query selectors in FAL, and how their effectiveness is related to the class balance of the queried data under varying levels of local heterogeneity ($\alpha$) and global class imbalance ($\rho$). 
Here, $\alpha$ denotes the concentration parameter of the widely used Dirichlet partition~\cite{mcmahan2017communication,li2020federated,karimireddy2020scaffold}, where smaller values correspond to stronger cross-client distribution shifts. 
The global imbalance ratio $\rho$ measures the long-tailed severity of the aggregated dataset~\cite{buda2018systematic,johnson2019survey}, with larger values indicating more skewed class frequencies.

Specifically, we conduct experiments on CIFAR-10~\cite{krizhevsky2009learning} with $\alpha {\in} \{0.1, 100\}$ and $\rho {\in} \{1, 20\}$\footnote{Local data distribution visualizations are shown in Appendix~\cref{sec:data_distribution}.}, evaluating both global and local models as query selectors using two representative AL strategies: entropy-based sampling~\cite{settles2009active} and coreset sampling~\cite{sener2017active}, representing uncertainty- and diversity-based querying, respectively.
At each AL round, 5\% of the training data is queried for labeling. We then evaluate the results from both quantitative and qualitative perspectives:


\begin{figure*}[t]
  \centering
   \includegraphics[width=0.97\linewidth]{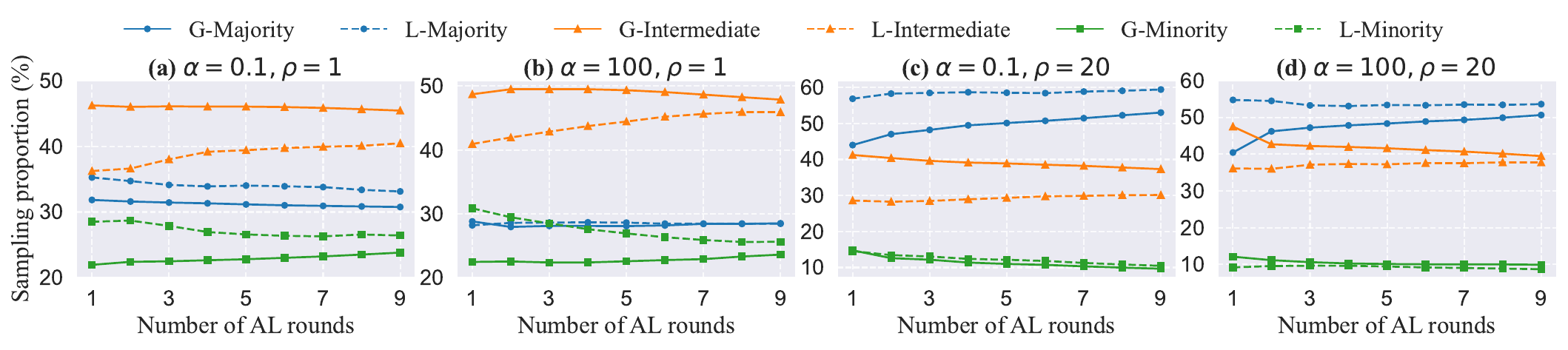}
   \caption{
   Cumulative sampling proportions of majority, intermediate, and minority classes for global (G) and local (L) query selectors under different $(\alpha,\rho)$ settings on CIFAR-10, where the three groups denote the top 3, middle 4, and bottom 3 classes by sample proportion.
}
   \label{fig:cumulative_ratio}
\vspace{-2mm}
\end{figure*}



\textbf{(1) Mean accuracy curves.}
Figure~\ref{fig:c10_curve} reports the mean test accuracy over $S{=}5$ random seeds.
These curves summarize the learning trajectory of each method, revealing its stability, convergence behavior, and whether it consistently maintains stronger performance throughout the AL process.

\textbf{(2) Robust paired analysis.}
To obtain a more aggregated and statistically grounded comparison, we compute the area under the learning curve (AULC) for each seed and conduct a paired analysis between any two methods $i$ and $j$ (Table~\ref{tab:entropy-coreset}).
For seed $s$, let $\Delta_s^{ij} = \text{AULC}_s^i - \text{AULC}_s^j$.  
We evaluate the set $\{\Delta_s^{ij}\}_{s=1}^S$ using three complementary statistics:
\begin{itemize}
  \item \textbf{Positive ratio ($\pi^+$)}~\cite{demvsar2006statistical} measures cross-seed consistency:
  \(
    \pi^+_{ij}
    = \frac{1}{S}\sum_{s=1}^S \mathbbm{1}(\Delta_s^{ij}>0).
  \)
  Larger values (e.g., $\ge 0.8$) indicate that $i$ consistently outperforms $j$.

  \item \textbf{One-sided Wilcoxon $p$-value}~\cite{wilcoxon1992individual} evaluates whether $i$ significantly outperforms $j$. 
  Small $p (\le 0.05)$ provides strong evidence favoring $i$, 
  $0.05<p\le0.10$ indicates a weak trend, 
  and $p>0.10$ suggests no difference.

  \item \textbf{Hodges--Lehmann (HL) estimator}~\cite{hodges2011estimates} provides a robust estimate of the typical performance gain:
  \[
    \widehat{\Delta}_{HL}^{ij}
    = \operatorname{median}\left\{
      \frac{\Delta_s^{ij}+\Delta_t^{ij}}{2} : 1 \le s \le t \le S
    \right\}.
  \] 
  Compared with the mean, HL is less sensitive to outliers and better reflects the gain in percentage points.
\end{itemize}

\begin{table}[t]
\centering
\small                             
\setlength{\tabcolsep}{1.5pt}        
\renewcommand{\arraystretch}{1.05} 
\begin{tabular}{c c c}
\toprule
\multirow{2}{*}{\diagbox[width=7.5em]{\ $(\alpha,\rho)$\ }{\ Method\ }}
& {\textbf{Entropy}}
& {\textbf{Coreset}} \\
\cmidrule(lr){2-3}
& \multicolumn{2}{c}{\ $(winner,\pi^+_{ij},p,\widehat{\Delta}_{HL}^{ij} )$\ } \\
\midrule
$(0.1, 1)$   & $(L, 1.0,0.03,201 \text{ pp})$ & $(G, 1.0,0.03,28 \text{ pp})$  \\
$(100, 1)$  & $(L, 1.0,0.03,50 \text{ pp})$ & $(G, 0.8,0.06,21 \text{ pp})$  \\
$(0.1, 20)$   & $(L, 1.0,0.03,66 \text{ pp})$ & $(G, 1.0,0.03,50 \text{ pp})$  \\
$(100, 20)$  & $(G, 1.0,0.03,106 \text{ pp})$ & $(G, 1.0,0.03,92 \text{ pp})$  \\
\bottomrule
\end{tabular}
\caption{
Pairwise comparison between global (G) and local (L) query selectors under different $(\alpha,\rho)$ configurations on CIFAR-10. 
Each entry reports the winner (strategy~$i$), positive ratio ($\pi^+_{ij}$), one-sided Wilcoxon $p$-value, and Hodges–Lehmann effect size $\widehat{\Delta}_{HL}^{ij}$ (in percentage points, pp) for Entropy and Coreset sampling.
}
\label{tab:entropy-coreset}
\vspace{-2mm}
\end{table}

\paragraph{Observation 1.} 
\textit{For uncertainty-based sampling, the local model generally outperforms the global model, except in cases where global class imbalance is severe while client distributions remain relatively homogeneous.}

As presented in Figure~\ref{fig:c10_curve} and Table~\ref{tab:entropy-coreset}, the local model consistently surpasses the global model under uncertainty-based querying across most $(\alpha,\rho)$ settings.
This trend is most pronounced when the global distribution is relatively balanced (small $\rho$) and client data are highly heterogeneous (small $\alpha$).
In this regime, although each client only considers its own data, the aggregation of diverse client-specific queries naturally produces a globally balanced query set. 
The global model, in contrast, tends to yield smoother and less discriminative uncertainty estimates due to cross-client averaging, making it less sensitive to subtle client-specific decision boundaries.
However, when global imbalance becomes severe (large $\rho$) and client distributions are nearly homogeneous (large $\alpha$), the diversity effect vanishes: local-model querying increasingly reflects the global long-tailed skew, leading to imbalanced acquisitions.
In such cases, the global model gains a noticeable advantage by exploiting shared knowledge across clients to counteract class skew during sampling.
These observations are further corroborated in Figure~\ref{fig:cumulative_ratio}, which visualizes how the cumulative sampling ratios of majority, intermediate, and minority classes evolve under different $(\alpha,\rho)$ configurations.

\paragraph{Observation 2.} 
\textit{The relative advantage of the local and global models is largely determined by which one achieves more class-balanced sampling—especially in acquiring minority classes—which closely aligns with final performance.}

Figures~\ref{fig:c10_curve} and~\ref{fig:cumulative_ratio} reveal a strong, consistent alignment between the balance of queried samples and downstream accuracy: models that acquire more minority-class instances invariably achieve higher performance. 
Under severe global class imbalance (large $\rho$), minority acquisition becomes the dominant factor. 
In Figure~\ref{fig:cumulative_ratio} (c), the global and local models obtain nearly the same number of minority samples in the initial AL rounds, yielding almost identical accuracy curves in Figure~\ref{fig:c10_curve} (c). 
As the gap in minority sampling grows and eventually stabilizes, so does the performance gap, suggesting a direct correspondence.
When the global distribution is balanced (small $\rho$), however, minority sampling alone no longer explains performance differences. 
In Figures~\ref{fig:cumulative_ratio} (a,b), both models acquire similar proportions of minority samples, yet the local model achieves clearly higher accuracy in Figure~\ref{fig:c10_curve} (a). 
This is because the local model’s queried set exhibits a more balanced class distribution, not only among minority classes but across the entire label space.


\paragraph{Observation 3.} 
\textit{For diversity-based sampling, the global model consistently outperforms the local model.}

Figures~\ref{fig:c10_curve} and Table~\ref{tab:entropy-coreset} show that, across all settings, the global model maintains a clear and consistent advantage when diversity-based strategies (i.e., Coreset sampling) are used.
Unlike uncertainty-driven querying, where performance is tied to the predictive scores, diversity-based sampling primarily relies on the structure of the feature space. Its effectiveness depends on how well feature embeddings capture global coverage, inter-sample separation, and manifold geometry.
Since the global model is optimized using aggregated updates from all clients, it develops more discriminative and globally aligned feature representations. 
These richer embeddings enable more faithful estimation of pairwise distances, allowing the global model to identify samples that better span the underlying data manifold.

\section{Methodology}


Based on the observations, we find that class-balanced sampling, especially for rare classes, is crucial for effective FAL. 
To this end, we propose FairFAL, an adaptive class-\textbf{Fair} \textbf{FAL} framework.
First, guided by \textbf{Observation~1}, we design an \emph{adaptive model-selection} mechanism that estimates both the severity of global class imbalance and the discrepancy between local and global class distributions on each client. This enables FairFAL to dynamically choose the most suitable query model for each client, either global or local.
Second, following \textbf{Observation~2}, we promote class-balanced querying through a \emph{prototype-guided pseudo-labeling} strategy that performs class-level sample selection. In particular, motivated by \textbf{Observation~3}, we exploit the global model’s well-generalized feature representations to compute per-class prototypes from labeled data on each client. The similarity between unlabeled samples and these prototypes is used to assign pseudo-labels, thereby guiding class-aware sample acquisition.
Finally, to further enhance query diversity, we introduce a \emph{two-stage balanced sampling} mechanism that jointly considers uncertainty and diversity. Concretely, we first construct a candidate pool by selecting high-uncertainty samples within each class, and then perform $k$-center sampling in a gradient-embedding space to obtain the final, diverse query set.

\subsection{Adaptive Model-Selection}
\label{subsec:adaptive_model_selection}

To adaptively decide whether the global or local model should act as the query selector, without introducing any additional privacy leakage, we estimate two key quantities on each client:
(1) the severity of global class imbalance, and 
(2) the divergence between local and global data distributions.  
These estimates, obtained solely from locally available labeled data and aggregated statistics, are then mapped to a continuous model-selection score that operationalizes the empirical observations in Section~\ref{sec:observation}.


Consider client $k$ with labeled set $\mathcal{D}_L^{(k)}$ and unlabeled pool $\mathcal{D}_U^{(k)}$.  
In the first AL round, labeled samples are typically acquired through random querying from $\mathcal{D}_U^{(k)}$, making $\mathcal{D}_L^{(k)}$ a reasonable IID approximation of the client’s underlying data distribution.  
We leverage this property to derive a privacy-preserving estimate of the global long-tailedness, which will later guide the adaptive model-selection process.


For client $k$, let $n_{k,c}$ denote the number of labeled samples of class $c$, and define 
\(
n_k^{\max} = \max_{c : n_{k,c} > 0} n_{k,c}.
\)
We construct a class-balanced subset
\begin{equation}
\mathcal{B}^{(k)}
= \bigcup_{c \in \mathcal{C}_k^+} \mathcal{B}^{(k)}_c,
\quad
\mathcal{B}^{(k)}_c \subseteq \mathcal{D}_L^{(k)}, 
\quad
|\mathcal{B}^{(k)}_c| = n_k^{\max},
\end{equation}
where $\mathcal{C}_k^+ = \{c\in\mathcal{C}: n_{k,c}>0\}$ and each $\mathcal{B}^{(k)}_c$ is formed by sampling (with replacement if necessary) from the labeled instances of class $c$. 
This upsampling step normalizes the per-class sample counts and ensures that every observed class contributes equally to the subsequent estimation.

\textbf{Global class imbalance ratio estimation.}
Let $f^{g}$ denote the current global model with parameters $\theta^{g}$, and let 
$\mathbf{p}^{g}(x)=f^{g}(x)$ be its predictive probability vector.
Using the class-balanced set $\mathcal{B}^{(k)}$, each client $k$ forms a local estimate of the global predictive prior:
\begin{equation}
\widehat{\boldsymbol{\pi}}^{(k)}_g
=\frac{1}{|\mathcal{B}^{(k)}|}
\sum_{x \in \mathcal{B}^{(k)}} \mathbf{p}^{g}(x).
\label{eq:local_prior}
\end{equation}
We estimate the global class imbalance ratio on client $k$ via
\begin{equation}
\gamma_k
= 
\frac{\min_{c \in \mathcal{C}_k^+} \widehat{\pi}_{g,c}}
     {\max_{c \in \mathcal{C}_k^+} \widehat{\pi}_{g,c}}
\in (0,1],
\end{equation}
where values closer to $1$ indicate a more balanced distribution, and smaller values reflect stronger long-tailed skew.

To ensure reliable yet privacy-preserving estimation, each client uploads only the scalar $\gamma_k$ to the server.  
The server aggregates these values via simple averaging,
\(
\bar{\gamma} = \frac{1}{K} \sum_{k=1}^K \gamma_k,
\)
and broadcasts $\bar{\gamma}$ to all clients as the global class-imbalance coefficient.  
Since labeled data in later AL rounds are no longer obtained through random sampling, 
$\bar{\gamma}$ is computed only in the first round and kept fixed thereafter.

\textbf{Local–global distributional divergence estimation.}
To characterize how heterogeneous each client is with respect to the global distribution, we further estimate the discrepancy between the local and global predictive priors.
Let $f^{(k)}$ denote the locally trained model on client $k$ with predictive vector $\mathbf{p}^{(k)}(x) = f^{(k)}(x)$.  
Using the same balanced set $\mathcal{B}^{(k)}$, client~$k$ computes
a local predictive prior $\widehat{\boldsymbol{\pi}}^{(k)}_\ell$.

We quantify the divergence between the global and local predictive priors by the normalized symmetric difference:
\begin{equation}
  d_k 
  = \frac{1}{C} \sum_{c=1}^C 
    \frac{\big|\widehat{\pi}_{g,c} - \widehat{\pi}^{(k)}_{\ell,c}\big|}
         {\widehat{\pi}_{g,c} + \widehat{\pi}^{(k)}_{\ell,c}},
  \label{eq:local_global_divergence}
\end{equation}
which satisfies $d_k \in [0,1]$ by construction.  
Larger values indicate stronger deviation of client~$k$ from the global predictive distribution, hence higher statistical heterogeneity.

Unlike the global class-balance coefficient $\bar{\gamma}$ estimated only once, $d_k$ is recomputed at every AL round to capture the evolving discrepancy between local and global labeled distributions as querying progresses.

\textbf{Continuous model-selection score.}
Section~\ref{sec:observation} indicates that the global model is more suitable when the global distribution is highly imbalanced and clients are relatively homogeneous, whereas local models dominate in balanced or highly heterogeneous regimes.

To capture this dependency in a smooth and data-driven manner, we define a continuous model-selection score $s_k \in [0,1)$ for each client $k$, based on the global balance coefficient $\bar{\gamma}$ and the local--global divergence $d_k$:
\begin{equation}
  s_k = 1 - \frac{1}{2}\big(d_k + \bar{\gamma}\big).
  \label{eq:score_simple}
\end{equation}
This score naturally recovers the four regimes discussed in Section~\ref{sec:observation}.
Larger $s_k$ values occur when the global distribution is highly imbalanced (small $\bar{\gamma}$) and the local distribution is well aligned with the global one (small $d_k$), corresponding to the regime where the global model is preferred.
In contrast, a more balanced global distribution or stronger local heterogeneity leads to smaller scores, favoring the local model. 
We then adopt a simple decision rule:
\begin{equation}
\label{eq:ms}
  \mathcal{M}^{(k)} =
  \begin{cases}
    \text{global model}, & s_k > \delta, \\
    \text{local model},  & s_k \le \delta,
  \end{cases}
\end{equation}
where $\delta$ is a fixed threshold (set to $0.75$).
This mechanism is lightweight and fully privacy-preserving, relying only on aggregated statistics derived from locally labeled data.


\subsection{Prototype-Guided Pseudo-Labeling}
\label{subsec:proto_pseudo}

As discussed in Section~\ref{sec:observation}, achieving class-balanced sampling is essential for improving FAL performance. 
While adaptive model selection determines which model should guide querying, it alone cannot guarantee balanced acquisition when class distributions are highly skewed. 
We therefore introduce a prototype-guided pseudo-labeling mechanism that explicitly promotes class-level balance.

In long-tailed or class-imbalanced settings, the classifier trained on imbalanced data tends to learn biased decision boundaries, making unlabeled samples more likely to be assigned to head classes. 
To counteract this bias, we replace direct classifier predictions with a prototype-based assignment: each class is represented by a feature prototype, and pseudo labels are determined by similarity between sample embeddings and these prototypes.

Motivated by \textbf{Observation~3}, which demonstrates that the global model provides more discriminative and well-generalized representations than local models, we adopt the global model as the feature extractor for prototype construction. 
This choice ensures that the resulting prototypes better reflect the global data manifold and yield more reliable class-wise pseudo labels.


Let $\phi^{g}(\cdot)$ denote the penultimate-layer feature extractor of the global model. 
On client $k$, given the labeled set $\mathcal{D}_L^{(k)}$, we first extract $\ell_2$-normalized features:
\begin{equation}
  \mathbf{z}_i^{(k)}
  = \frac{\phi^{g}(x_i)}{\|\phi^{g}(x_i)\|_2},
  \quad (x_i, y_i) \in \mathcal{D}_L^{(k)}.
\end{equation}

For each class $c \in \mathcal{C}_k^+$, we compute a class prototype by averaging the normalized features of its labeled samples:
\begin{equation}
  \boldsymbol{\mu}_c^{(k)}
  = \frac{1}{|\mathcal{D}_{L,c}^{(k)}|}
    \sum_{(x_i, y_i)\in\mathcal{D}_L^{(k)},\, y_i=c}
    \mathbf{z}_i^{(k)},
  \label{eq:proto_def}
\end{equation}
where $\mathcal{D}_{L,c}^{(k)}$ denotes the subset of labeled instances of class $c$ at client $k$.
These prototypes provide a compact, noise-reduced summary of each class in the representation space.

For any unlabeled sample $x \in \mathcal{D}_U^{(k)}$, we extract its normalized feature $\mathbf{z}^{(k)}(x)$ and compute cosine similarity to all available prototypes:
\begin{equation}
  s_c^{(k)}(x)
  = \left\langle \mathbf{z}^{(k)}(x), \, \boldsymbol{\mu}_c^{(k)} \right\rangle,
  \quad c \in \mathcal{C}_k^+.
\end{equation}
Then, a prototype-guided pseudo-label is assigned via
\begin{equation}
  \widehat{y}^{(k)}(x)
  = \arg\max_{c \in \mathcal{C}_k^+} s_c^{(k)}(x),
  \label{eq:pseudo_label}
\end{equation}
which leverages the geometry of the global feature space rather than raw classifier logits, offering improved robustness under long-tailed and heterogeneous federation.

The resulting pseudo labels induce a partition of the unlabeled pool into class-specific subsets:
\begin{equation}
  \widetilde{\mathcal{D}}_{U,c}^{(k)}
  = \left\{\, x \in \mathcal{D}_U^{(k)} \,\middle|\, \widehat{y}^{(k)}(x) = c \,\right\},
  \quad c \in \mathcal{C}_k^+.
\end{equation}
These subsets form the basis of the class-aware sampling mechanism in the subsequent stage.

\subsection{Uncertainty--Diversity Balanced Sampling}
\label{subsec:uncertainty_diversity}
While class-wise sampling promotes balanced label acquisition, directly selecting the most uncertain samples within each class often results in severe redundancy: highly uncertain examples tend to cluster in a narrow region of the feature space, wasting annotation budget on near-duplicates. To address this, we introduce a two-stage sampling strategy that jointly enforces uncertainty and diversity.

\textbf{Stage~1: Class-wise candidate selection.}
For client $k$, let $b_c^{(k)}$ denote the uniformly allocated query budget for class $c$, with $\sum_c b_c^{(k)} = B_k$.
Using the adaptively selected query model $\mathcal{M}^{(k)}$, we compute an uncertainty score (e.g., entropy) $u^{(k)}(x)$ for each $x \in \widetilde{\mathcal{D}}_{U,c}^{(k)}$. 
We then construct an over-complete candidate pool:
\begin{equation}
  \mathcal{H}_c^{(k)}
  = \operatorname{TopK}\!\Big(
      \widetilde{\mathcal{D}}_{U,c}^{(k)},\;
      u^{(k)}(x),\;
      \kappa\, b_c^{(k)}
    \Big),\quad \kappa>1,
  \label{eq:candidate_pool}
\end{equation}
which retains the $\kappa b_c^{(k)}$ most uncertain samples for class $c$.  
This stage preserves informativeness while maintaining sufficient redundancy for diversity refinement.

\textbf{Stage~2: Diversity sampling.}
To reduce redundancy within each $\mathcal{H}_c^{(k)}$, we refine candidates in a global gradient-embedding space consistently using the global model.

We compute the gradient embedding for any $x$ as
\begin{equation}
  \mathbf{g}^{(k)}(x)
  = \psi\!\big(x;\phi^{g},f^{g}\big) \in \mathbb{R}^d,
  \label{eq:grad_embed}
\end{equation}
where $\psi(\cdot)$ denotes the gradient of the classification loss with respect to the global classifier parameters, using global features $\phi^{g}(x)$. This gradient embedding captures both semantic structure and class-discriminative geometry, providing a stable and globally aligned space for enforcing diversity—particularly beneficial for rare classes.

\begin{figure*}[!ht]
\centering
   \includegraphics[width=1\linewidth]{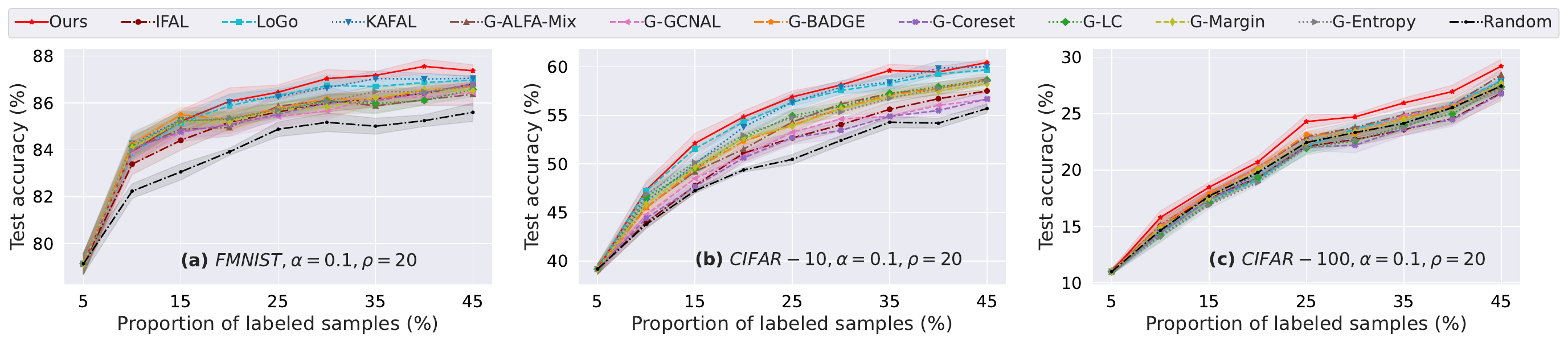}
   \includegraphics[width=1\linewidth]{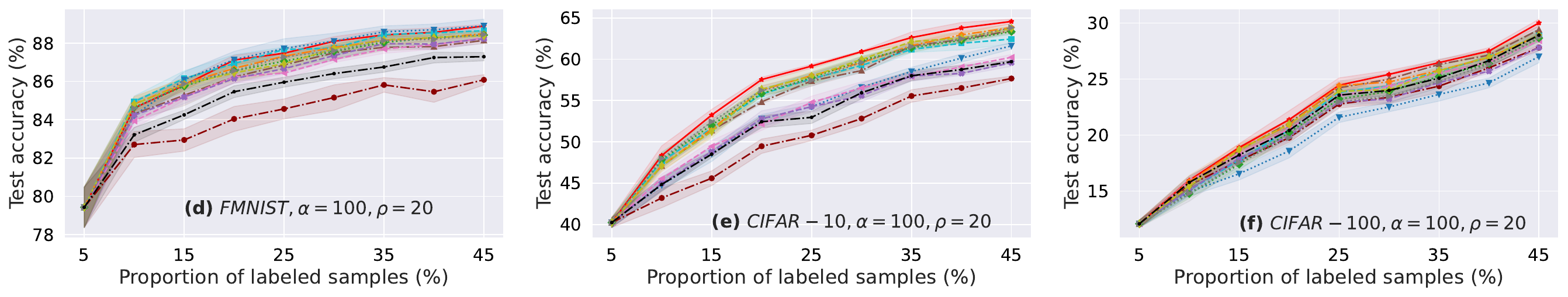}
   \includegraphics[width=1\linewidth]{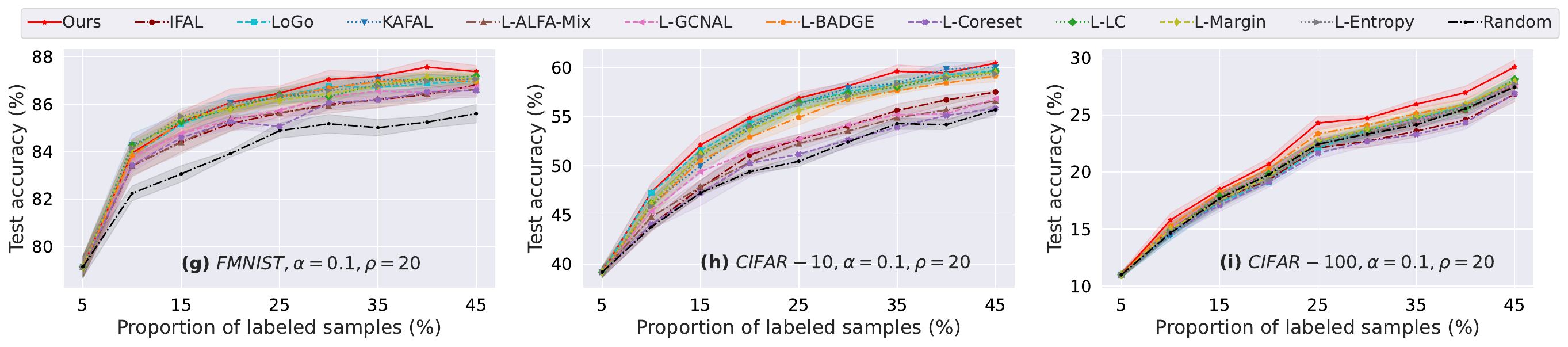}
   \includegraphics[width=1\linewidth]{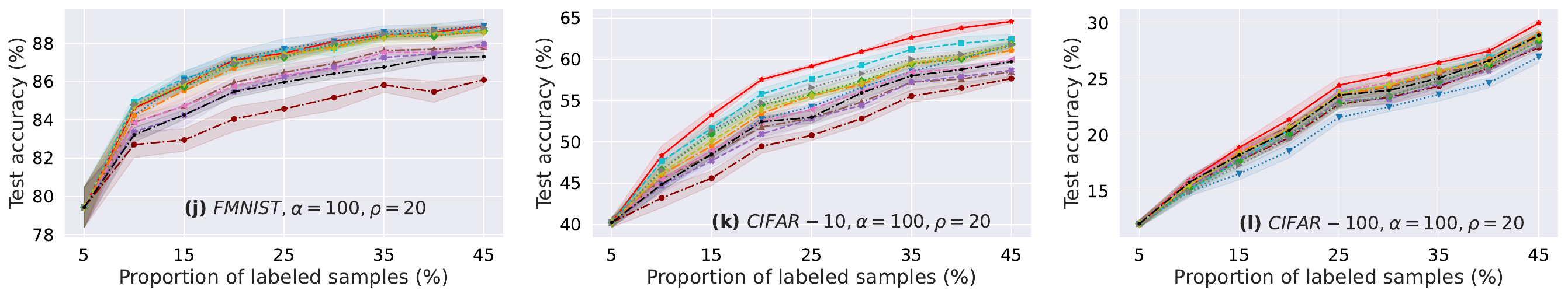}
\caption{Test accuracy vs.\ proportion of labeled samples on FMNIST, CIFAR-10, and CIFAR-100 under global imbalance $\rho=20$ and two heterogeneity levels ($\alpha=0.1$ and $\alpha=100$). 
Each subplot corresponds to a dataset–heterogeneity setting, and curves compare FairFAL with FAL and AL baselines using global (G-) or local (L-) models. Numerical results are provided in Tables~\ref{tab:main_numeric} and~\ref{tab:main_numeric2} in the Appendix. }
\vspace{-4mm}
	\label{fig:main_results}
\end{figure*}

For each class $c$, we define the embedding sets for labeled (anchors) and candidate samples as:
\begin{align}
  \mathcal{A}_c^{(k)}
  &= \{\mathbf{g}^{(k)}(x) : (x,y)\in\mathcal{D}_L^{(k)},\, y=c\}, \\
  \mathcal{G}_c^{(k)}
  &= \{\mathbf{g}^{(k)}(x) : x\in\mathcal{H}_c^{(k)}\}.
\end{align}
Treating $\mathcal{A}_c^{(k)}$ as fixed anchors, we select $b_c^{(k)}$ additional points from $\mathcal{H}_c^{(k)}$ such that the combined set achieves good coverage in the embedding space.  
This is formulated as a class-wise $k$-center objective:
\begin{equation}
  \mathcal{Q}_c^{(k)}
  = \argmin_{\substack{\mathcal{Q}'\subset\mathcal{H}_c^{(k)}\\|\mathcal{Q}'|=b_c^{(k)}}}
      \Bigg\{
        \max_{x\in\mathcal{H}_c^{(k)}}
        \min_{a\in\mathcal{A}_c^{(k)}\cup\mathcal{Q}'}
        d\!\big(\mathbf{g}^{(k)}(x), \mathbf{g}^{(k)}(a)\big)
      \Bigg\}.
  \label{eq:kcenter_obj}
\end{equation}
However, Eq. \eqref{eq:kcenter_obj} is NP-hard~\cite{sener2017active}, so we adopt the greedy $k$-center algorithm~\cite{sener2017active} to get an efficient $2$-approximation.



\textbf{Final query set.}
The final query set for client $k$ is the union of class-wise selections:
\begin{equation}
  \mathcal{Q}^{(k)} 
  = \bigcup_{c\in\mathcal{C}_k^+} \mathcal{Q}_c^{(k)}.
\end{equation}

Overall, the two-stage procedure first identifies informative samples based on uncertainty and then promotes representative coverage through diversity under class-specific budget constraints.  
This yields query sets that are simultaneously informative, diverse, and class-balanced, enabling more sample-efficient and fairer FAL.
\section{Experiments}

\begin{table*}[t]
\centering
\small
\setlength{\tabcolsep}{8pt}
\renewcommand{\arraystretch}{1.15}
\begin{tabular}{l cc cc cccc}
\toprule
\multirow{2}{*}{\diagbox[width=3.cm]{$(\alpha,\rho)$}{Component}}
  & \multicolumn{2}{c}{w/ Model selection}
  & \multicolumn{2}{c}{w/ Class-wise sampling}
  & \multicolumn{4}{c}{w/ Two stage balanced sampling}\\
\cmidrule(lr){2-3} \cmidrule(lr){4-5} \cmidrule(lr){6-9}
& $\mathcal{M}^{(k)}$ & $\tilde{\mathcal{M}}^{(k)}$ & Local & Global & $\kappa=2$ & $\kappa=3$ & $\kappa=4$ \textbf{(Final)} & $\kappa=5$ \\
\midrule
$(0.1, 20)$   & 59.33  & 58.49  & 59.14 & 59.95  & 60.61  & 60.38  & 60.44  & 60.28  \\
$(100, 20)$   & 63.65  & 61.89  & 63.39 & 64.02   & 64.60  & 64.58  & 64.57  & 64.17  \\
\bottomrule
\end{tabular}
\vspace{-1mm}
\caption{Ablation results of FairFAL under different $(\alpha,\rho)$ configurations. We evaluate three key components: adaptive model selection, prototype-guided class-wise sampling, and the two-stage uncertainty–diversity balanced sampling strategy. Modules are added sequentially from left to right, and the final configuration corresponds to $\kappa{=}4$.
All results are reported in terms of final-round test accuracy (\%).
}
\vspace{-3mm}
\label{tab:ablation}
\end{table*}

\textbf{Datasets.}
We evaluate FairFAL on five benchmarks spanning natural and medical image classification. 
The natural-image datasets include FMNIST~\cite{xiao2017fashion}, CIFAR-10~\cite{krizhevsky2009learning}, and CIFAR-100~\cite{krizhevsky2009learning}, while the medical benchmarks comprise OctMNIST and DermaMNIST from MedMNIST~\cite{yang2023medmnist}. 
To highlight the effect of global class imbalance, we set the overall imbalance ratio to $\rho{=}20$ in all experiments. 
Data are then partitioned across 10 clients using a Dirichlet distribution with $\alpha{=}0.1$ and $100$, representing highly heterogeneous and nearly homogeneous client splits, respectively.


\begin{table}[t]
\centering
\small
\setlength{\tabcolsep}{4.pt}
\renewcommand{\arraystretch}{1.1}
\begin{tabular}{l| c| c c c| c}
\toprule
& Random 
& KAFAL
& LoGo
& IFAL
& \textbf{Ours} \\
\midrule
OctMNIST
& 68.30 & 70.40 & 70.00 & 68.40 & \textbf{72.80} \\
DermaMNIST
& 72.32 & 73.27 & 73.62 & 72.97 & \textbf{73.77} \\
\bottomrule
\end{tabular}
\vspace{-1mm}
\caption{Final-round test accuracy comparison on OctMNIST and DermaMNIST under extreme client heterogeneity $\alpha{=}0.1$.}
\label{tab:medical_results}
\vspace{-3mm}
\end{table}


\textbf{Baselines.}
We compare FairFAL against 11 query strategies grouped into five categories: 
(1) Random sampling; 
(2) Uncertainty-based methods, including Entropy~\cite{holub2008entropy}, Margin~\cite{balcan2007margin}, and Least Confidence (LC)~\cite{li2006confidence}; 
(3) Diversity-based method Coreset~\cite{sener2017active}; 
(4) Hybrid approaches such as BADGE~\cite{ash2019deep}, GCNAL~\cite{caramalau2021sequential}, and ALFA-Mix~\cite{parvaneh2022active}, which jointly leverage uncertainty and diversity cues; 
(5) FAL methods, including KAFAL~\cite{cao2023knowledge}, LoGo~\cite{kim2023re}, and IFAL~\cite{ijcai2025p812}. 

We provide training configurations, dataset details, and additional experimental results in Appendix~\cref{sec:training_details}.

\subsection{Performance Comparison}

Figure~\ref{fig:main_results} compares test accuracy across AL rounds on FMNIST, CIFAR-10, and CIFAR-100 under severe global class imbalance and both heterogeneous and homogeneous client distributions.
FairFAL consistently outperforms all baselines, with its advantage becoming more pronounced as task difficulty increases. The widening gap from FMNIST to CIFAR-100 highlights its robustness and generalization in challenging federated settings.
In contrast, the gains of prior AL and FAL methods over Random sampling diminish as the task becomes harder, especially under nearly homogeneous client distributions ($\alpha{=}100$). Under high heterogeneity ($\alpha{=}0.1$), client-specific distributions naturally inject diversity into the global query set, allowing many baselines to perform reasonably well. However, when this implicit diversity weakens, methods such as IFAL, which lack explicit class-balancing mechanisms, tend to query redundant or biased samples, causing their performance to approach or even fall below that of random sampling.
This further highlights the necessity and effectiveness of FairFAL’s strategy design, where adaptive model selection works jointly with class-balanced, diversity-aware sampling.


Table~\ref{tab:medical_results} reports results on two naturally imbalanced medical benchmarks, OctMNIST and DermaMNIST, under severe client heterogeneity ($\alpha{=}0.1$). Unlike the earlier curated datasets, both benchmarks exhibit intrinsic long-tailed label distributions, posing a substantially more challenging setting for class-fair querying. Despite this difficulty, FairFAL consistently achieves the highest accuracy. These results further underscore the robustness of FairFAL in highly imbalanced, real-world clinical scenarios.

\subsection{Ablation Studies}
Table~\ref{tab:ablation} summarizes the ablation results under two representative configurations, where the components of FairFAL are incrementally enabled from left to right.

\textbf{Effectiveness of adaptive model selection.}
We denote by $\mathcal{M}^{(k)}$ the model (global or local) chosen by our adaptive rule on client~$k$, and by $\widetilde{\mathcal{M}}^{(k)}$ its complementary option, i.e., the model not selected. 
Across both heterogeneity levels, using $\mathcal{M}^{(k)}$ consistently leads to higher accuracy, confirming that adaptively selecting the appropriate query model is crucial for stabilizing querying behavior in FAL.


\textbf{Role of global vs.\ local prototypes.}
To assess the effect of prototype quality on class-wise sampling, we compare prototypes constructed using features from the local and global models. In both cases, prototypes based on global features perform better, suggesting that globally aggregated representations yield cleaner class separation and more reliable similarity estimates, thereby enabling more accurate pseudo-labels and more balanced class-aware acquisition.


\textbf{Effect of candidate-pool size.}
We then ablate the over-sampling ratio $\kappa$ used to construct each class-specific candidate pool. 
The two-stage strategy consistently outperforms its one-stage counterparts across different choices of $\kappa$. Although $\kappa{=}2$ achieves the highest accuracy by a small margin, the differences among $\kappa{=}2$, $3$, and $4$ are minor, suggesting that the method is relatively insensitive to the exact choice of $\kappa$.
We therefore adopt $\kappa{=}4$ as FairFAL's final configuration, as it offers a larger and more flexible candidate pool for subsequent diversity refinement.

\textbf{Additional Appendix results} further verify FairFAL’s robustness across different FL frameworks, client numbers, model architectures, uncertainty measures, and $\delta$ values.

\section{Conclusion}

In this paper, we revisited FAL under realistic challenges from global long-tailed distributions and client heterogeneity. Our systematic analysis shows that class-balanced sampling is a key factor in FAL performance, and that the choice between global and local query selectors depends jointly on global imbalance and local–global divergence. Building on these findings, we propose FairFAL, a class-fair FAL framework that integrates adaptive model selection, prototype-guided pseudo-labeling, and uncertainty–diversity balanced sampling. FairFAL consistently improves sample efficiency and robustness across diverse datasets and settings, showing strong effectiveness in extreme non-IID and severely imbalanced federated scenarios.

\section*{Acknowledgements}
This work was supported in part by the NSFC (U2441285).

{
    \small
    \bibliographystyle{ieeenat_fullname}
    \bibliography{main}
}

\appendix
\clearpage
\setcounter{page}{1}
\maketitlesupplementary

\section{Visualization of Local Data Distributions}
\label{sec:data_distribution}

\begin{figure*}[!ht]
\centering
   \includegraphics[width=0.46\linewidth]{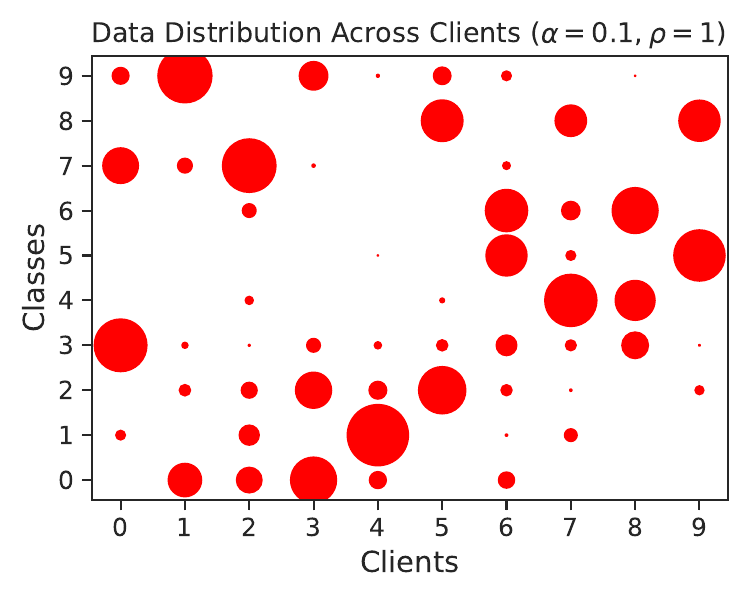}
   \includegraphics[width=0.46\linewidth]{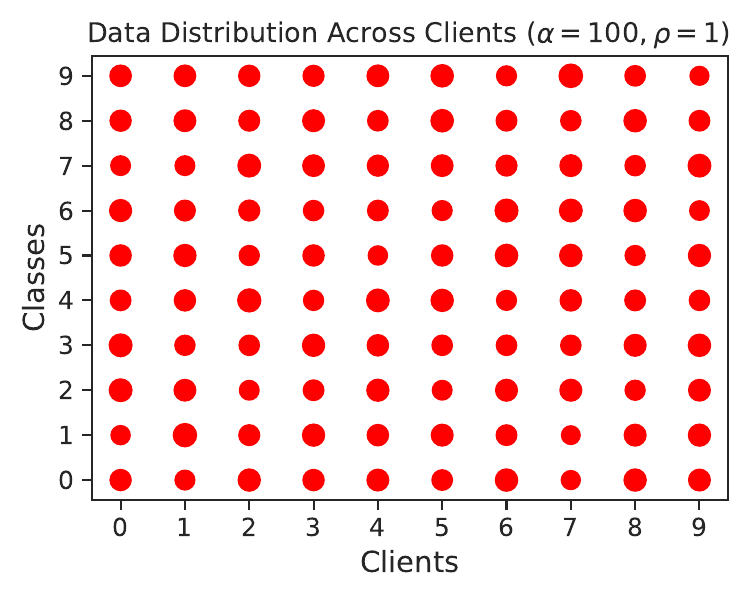}
   \includegraphics[width=0.46\linewidth]{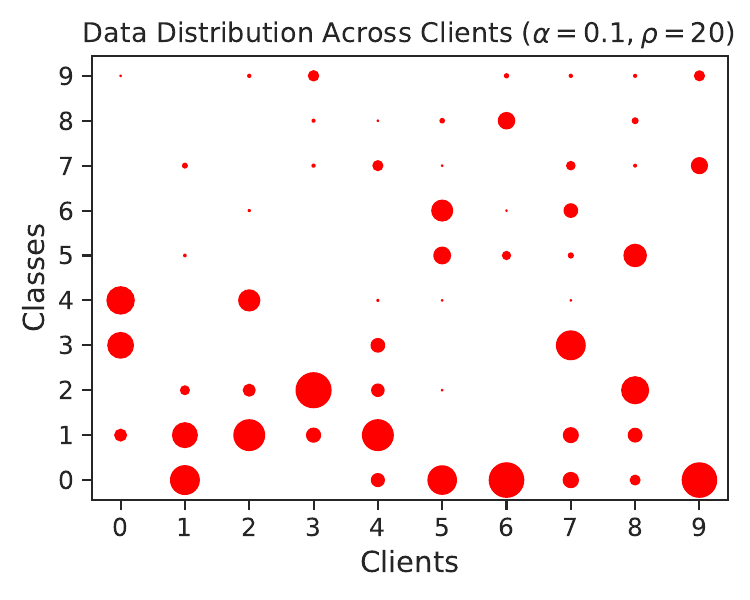}
   \includegraphics[width=0.46\linewidth]{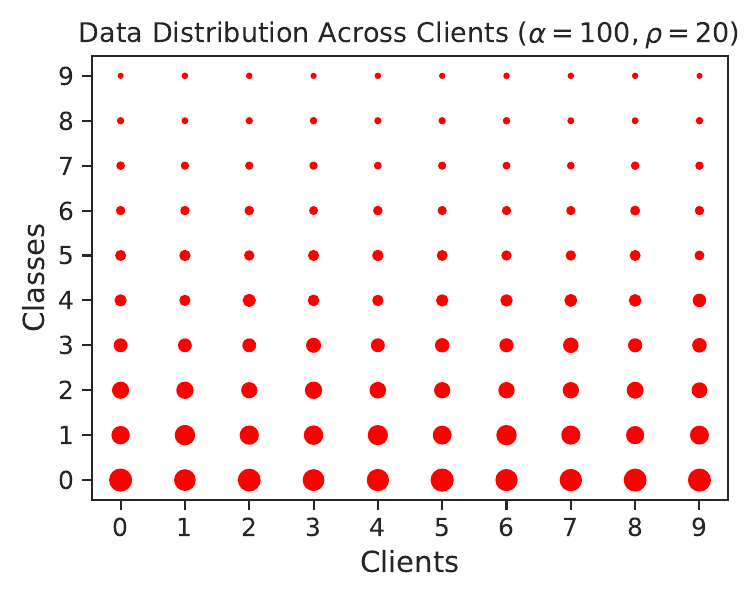}
\caption{Visualization of the data distributions under the four $(\alpha,\rho)$ configurations discussed in the Observation section. These plots illustrate how client heterogeneity and global class imbalance jointly shape the federated data landscape.}
	\label{fig:dd}
\end{figure*}

To illustrate the structure of the federated data, Figure~\ref{fig:dd} visualizes the client-wise class distributions under four representative $(\alpha,\rho)$ configurations.
The Dirichlet parameter $\alpha$ controls local heterogeneity: smaller values produce highly skewed per-client distributions, whereas larger values yield more uniform splits.
In contrast, the global imbalance ratio $\rho$ determines the global long-tailedness: small $\rho$ indicates a nearly balanced overall distribution, while large $\rho$ induces pronounced head–tail patterns across classes.
Together, these examples highlight how different combinations of $\alpha$ and $\rho$ shape the federated landscape and motivate the need for class-fair sampling in federated active learning.

\section{Additional Experimental Details}
\label{sec:training_details}

\subsection{Training Details}
All experiments are carried out under the standard FedAvg framework~\cite{mcmahan2017communication}. 
We train for $100$ communication rounds, with each client performing $5$ local epochs per round. 
FAL proceeds for $9$ query cycles: in the first cycle, each client randomly labels $5\%$ of its local data to initialize $\mathcal{D}_L^{(k)}$, and each subsequent cycle queries an additional $5\%$ from its unlabeled pool.
We adopt a 4-layer CNN backbone and optimize it with SGD (momentum $0.9$, weight decay $1{\times}10^{-5}$, batch size $64$). 
The learning rate is set to $0.01$ and decayed by a factor of $10$ after round $75$. 
For our method, the over-sampling ratio $\kappa$ in Eq.~\eqref{eq:candidate_pool} is fixed to $4$.
All experiments are conducted over five random seeds $(seed=1,2,3,4,5)$ on NVIDIA GeForce RTX~3090 GPUs, and we report the mean accuracy across runs.

\subsection{Dataset Details}

\begin{table*}[t]
\centering
\begin{tabular}{lccc}
\toprule
\textbf{Dataset} & \textbf{\#Classes} & \textbf{Train / Test} & \textbf{Imbalance} \\
\midrule
FMNIST~\cite{xiao2017fashion}
& 10 
& 60k / 10k 
& Balanced \\
CIFAR-10~\cite{krizhevsky2009learning}
& 10 
& 50k / 10k 
& Balanced \\
CIFAR-100~\cite{krizhevsky2009learning}
& 100 
& 50k / 10k 
& Balanced \\
OCTMNIST~\cite{yang2023medmnist}
& 4 
& 97{,}477 / 10{,}832
& Long-tailed \\
DermaMNIST~\cite{yang2023medmnist}
& 7 
& 7{,}007 / 2{,}003
& Long-tailed \\
\bottomrule
\end{tabular}
\caption{Summary of datasets used in our experiments, including natural-image (FMNIST, CIFAR-10, and CIFAR-100) and medical-image benchmarks (OCTMNIST and DermaMNIST). 
Long-tailed indicates naturally imbalanced label distributions.}
\label{tab:dataset_summary}
\end{table*}

\begin{table*}[t]
\centering
\setlength{\tabcolsep}{4.5pt}
\renewcommand{\arraystretch}{1.15}
\begin{tabular}{l | c | cccc | cccc | cccc }
\toprule
\multirow{2}{*}{\textbf{Method}} & \multirow{2}{*}{\textbf{Model}} &
\multicolumn{4}{c|}{\textbf{FMNIST}} &
\multicolumn{4}{c|}{\textbf{CIFAR-10}} &
\multicolumn{4}{c}{\textbf{CIFAR-100}} \\
& & 15\% & 25\% & 35\% & 45\% & 15\% & 25\% & 35\% & 45\% & 15\% & 25\% & 35\% & 45\% \\
\midrule
Random      & -- & 83.06 & 84.88 & 85.01 & 85.60 
& 47.24 & 50.46 & 54.29 & 55.70 
& 17.69 & 22.44 & 24.13 & 27.44 \\
\midrule
\multirow{2}{*}{Entropy~\cite{holub2008entropy}} & $G$ &
85.20 & 85.71 & 86.48 & 86.72 &
50.15 & 54.79 & 56.76 & 58.54 &
16.89 & 22.01 & 23.74 & 27.15  \\ 
& $L$ &
\textbf{85.49} & 86.36 & 86.76 & 87.06 &
50.93 & 56.26 & 58.34 & 59.35 &
17.44 & 22.52 & 24.56 & 27.70  \\ 
\midrule
\multirow{2}{*}{Margin~\cite{balcan2007margin}} & $G$ &
85.21 & 85.71 & 86.23 & 86.53 &
49.66 & 53.90 & 56.89 & 58.24 &
17.52 & 22.46 & 24.47 & 27.70  \\
& $L$ &
\underline{85.43} & 86.16 & 86.76 & 87.04 &
51.16 & 55.64 & 58.28 & 59.49 &
17.51 & 22.71 & 24.80 & 27.92  \\ 
\midrule
\multirow{2}{*}{LC~\cite{li2006confidence}} & $G$ &
85.24 & 85.69 & 85.97 & 86.57 &
49.54 & 54.95 & 57.27 & 58.61 &
17.03 & 21.93 & 23.94 & 27.37  \\ 
& $L$ &
85.26 & \underline{86.37} & 86.82 & \underline{87.19} &
51.20 & \underline{56.44} & 58.01 & 59.62 &
17.88 & 22.66 & 24.72 & 28.13  \\  
\midrule
\multirow{2}{*}{Coreset~\cite{sener2017active}} & $G$ &
84.77 & 85.52 & 86.09 & 86.80 &
47.66 & 52.69 & 54.89 & 56.69 &
17.36 & 22.03 & 23.65 & 26.75  \\
& $L$ &
84.58 & 85.06 & 86.18 & 86.57 &
47.23 & 51.17 & 53.92 & 55.82 &
17.08 & 21.66 & 23.30 & 26.87  \\ 
\midrule
\multirow{2}{*}{BADGE~\cite{ash2019deep}} & $G$ &
\textbf{85.49} & 85.77 & 86.13 & 86.70 &
49.43 & 53.98 & 57.10 & 58.68 &
17.92 & 23.16 & 24.62 & 27.37  \\ 
& $L$ &
85.25 & 86.30 & 86.95 & 86.94 &
50.53 & 54.93 & 57.65 & 59.13 &
\underline{18.20} & \underline{23.36} & \underline{25.12} & 27.95  \\ 
\midrule
\multirow{2}{*}{GCNAL~\cite{caramalau2021sequential}} & $G$ &
84.73 & 85.42 & 86.25 & 86.44 &
48.50 & 53.26 & 54.89 & 56.65 &
17.95 & 22.51 & 24.79 & 27.56  \\ 
& $L$ &
84.75 & 85.73 & 86.55 & 86.74 &
49.38 & 52.73 & 55.20 & 56.83 &
17.89 & 22.82 & 24.53 & 27.70  \\ 
\midrule
\multirow{2}{*}{ALFA-Mix~\cite{parvaneh2022active}} & $G$ &
84.88 & 85.86 & 85.88 & 86.37 &
49.20 & 54.31 & 57.27 & 58.75 &
17.86 & 22.95 & 24.90 & \underline{28.43} \\ 
& $L$ &
84.38 & 85.63 & 86.22 & 86.61 &
47.87 & 52.27 & 54.90 & 56.62 &
17.73 & 22.39 & 24.35 & 27.58  \\
\midrule
KAFAL~\cite{cao2023knowledge} & $G,L$ &
85.27 & 86.27 & \underline{87.04} & 87.05 &
49.99 & 56.34 & \underline{58.41} & \underline{60.01} &
17.36 & 22.52 & 24.46 & 27.84 \\
\midrule
LoGo~\cite{kim2023re} & $G,L$ &
85.15 & 86.31 & 86.70 & 86.98 &
\underline{51.56} & 56.35 & 58.30 & 59.68 &
17.33 & 22.08 & 24.74 & 27.95 \\
\midrule
IFAL~\cite{ijcai2025p812} & $G,L$ &
84.41 & 85.62 & 86.18 & 86.80 &
47.76 & 52.67 & 55.62 & 57.51 &
17.56 & 22.10 & 23.58 & 26.82 \\
\midrule
\textbf{FairFAL (ours)} & $G,L$ &
85.21 & \textbf{86.46} & \textbf{87.18} & \textbf{87.37} &
\textbf{52.12} & \textbf{56.90} & \textbf{59.62} & \textbf{60.44} &
\textbf{18.47} & \textbf{24.29} & \textbf{25.94} & \textbf{29.20}  \\
\bottomrule
\end{tabular}
\caption{Test accuracy (\%) comparison on three benchmark datasets under different labeled-data ratios, with $\alpha = 0.1$ and $\rho = 20$. Results are averaged over five random seeds. Traditional AL baselines are evaluated using two query-selector models, $G$ and $L$, denoting global and local selectors, respectively. \textbf{Bold} numbers indicate the best result in each column, while \underline{underlined} numbers denote the second best.
}
\label{tab:main_numeric}
\end{table*}

\begin{itemize}
    \item \textbf{Fashion-MNIST (FMNIST)}~\cite{xiao2017fashion} is a gray-scale image dataset of fashion items, containing $60{,}000$ training and $10{,}000$ test images at a resolution of $28{\times}28$. 
    The dataset spans $10$ clothing categories (e.g., T-shirt, trouser, coat, sneaker) and is widely used as a lightweight replacement for MNIST in image classification benchmarks.

    \item \textbf{CIFAR-10}~\cite{krizhevsky2009learning} contains $60{,}000$ natural images of size $32{\times}32$ with $3$ color channels, split into $50{,}000$ training and $10{,}000$ test samples.
    The dataset covers $10$ object classes (such as plane, car, bird, and dog) with balanced class frequencies, and serves as a standard benchmark for supervised and active learning.

    \item \textbf{CIFAR-100}~\cite{krizhevsky2009learning} shares the same image resolution and train/test split as CIFAR-10 but includes $100$ fine-grained categories, each with $500$ training and $100$ test images.
    Compared to CIFAR-10, it presents a more challenging multi-class classification task due to its larger label space and higher intra-class variability.

    \item \textbf{OCTMNIST}~\cite{yang2023medmnist} is a medical imaging dataset derived from retinal optical coherence tomography (OCT) scans for retinal disease diagnosis.
    It comprises $109{,}309$ gray-scale images and is framed as a $4$-class classification task, with all images center-cropped and resized to $1{\times}28{\times}28$. 
    Following the official MedMNIST protocol, the dataset is split into training, validation, and test sets. Its label distribution is naturally long-tailed, mirroring the real-world prevalence imbalance among retinal conditions.

    \item \textbf{DermaMNIST}~\cite{yang2023medmnist} is built from the HAM10000 collection of dermatoscopic skin-lesion images.
    It contains $10{,}015$ color images spanning $7$ diagnostic categories, resized to $3{\times}28{\times}28$ and split into training, validation, and test sets following a $7{:}1{:}2$ ratio. 
    The dataset exhibits a pronounced long-tailed distribution (e.g., malignant cases are significantly rarer than benign ones), making it a representative benchmark for evaluating models under class-imbalanced medical image classification.

\end{itemize}

\subsection{Numerical Results of Figure 3}

\begin{table*}[t]
\centering
\setlength{\tabcolsep}{4.5pt}
\renewcommand{\arraystretch}{1.15}
\begin{tabular}{l | c | cccc | cccc | cccc }
\toprule
\multirow{2}{*}{\textbf{Method}} & \multirow{2}{*}{\textbf{Model}} &
\multicolumn{4}{c|}{\textbf{FMNIST}} &
\multicolumn{4}{c|}{\textbf{CIFAR-10}} &
\multicolumn{4}{c}{\textbf{CIFAR-100}} \\
& & 15\% & 25\% & 35\% & 45\% & 15\% & 25\% & 35\% & 45\% & 15\% & 25\% & 35\% & 45\% \\
\midrule
Random      & -- & 84.26 & 85.96 & 86.76 & 87.29
& 48.51 & 52.96 & 58.00 & 59.70 
& 18.23 & 23.57 & 25.09 & 28.93 \\
\midrule
\multirow{2}{*}{Entropy~\cite{holub2008entropy}} & $G$ &
85.88 & 87.28 & 88.14 & 88.46 &
\underline{52.43} & 57.47 & 61.53 & \underline{63.83} &
17.56 & 22.98 & 24.80 & 28.77  \\ 
& $L$ &
85.98 & 87.34 & \underline{88.50} & \textbf{88.80} &
51.37 & 56.59 & 60.01 & 61.74 &
17.33 & 22.66 & 24.88 & 27.97  \\ 
\midrule
\multirow{2}{*}{Margin~\cite{balcan2007margin}} & $G$ &
85.84 & 86.92 & 88.18 & 88.38 &
51.26 & 57.93 & \underline{62.13} & 63.72 &
18.68 & 23.92 & 25.76 & 28.91  \\
& $L$ &
85.89 & 87.39 & 88.29 & 88.61 &
50.09 & 55.59 & 59.58 & 61.68 &
18.27 & 23.71 & 25.78 & 28.70  \\ 
\midrule
\multirow{2}{*}{LC~\cite{li2006confidence}} & $G$ &
85.75 & 87.06 & 88.04 & 88.43 &
52.04 & 57.78 & 61.38 & 63.33 &
17.39 & 23.22 & 25.23 & 28.66  \\ 
& $L$ &
85.71 & 87.25 & 88.33 & 88.64 &
50.96 & 55.73 & 59.43 & 61.80 &
17.64 & 22.85 & 24.80 & 28.38  \\  
\midrule
\multirow{2}{*}{Coreset~\cite{sener2017active}} & $G$ &
85.18 & 86.67 & 87.98 & 88.23 &
48.81 & 54.26 & 57.97 & 59.57 &
17.93 & 22.95 & 24.70 & 27.82  \\
& $L$ &
84.24 & 86.24 & 87.25 & 87.96 &
47.71 & 52.81 & 57.24 & 58.66 &
17.87 & 23.08 & 24.56 & 28.04  \\ 
\midrule
\multirow{2}{*}{BADGE~\cite{ash2019deep}} & $G$ &
85.89 & 87.32 & 88.15 & 88.50 &
51.54 & \underline{58.02} & 61.65 & \underline{63.83} &
\underline{18.76} & \underline{24.42} & 25.78 & 28.91  \\ 
& $L$ &
85.51 & 87.43 & 88.40 & 88.58 &
49.52 & 55.59 & 59.42 & 61.06 &
18.47 & 23.53 & 25.65 & 29.05  \\ 
\midrule
\multirow{2}{*}{GCNAL~\cite{caramalau2021sequential}} & $G$ &
85.18 & 86.44 & 87.70 & 88.27 &
49.46 & 54.77 & 57.90 & 60.23 &
18.34 & 23.47 & 25.54 & 28.42  \\ 
& $L$ &
84.73 & 86.33 & 87.53 & 87.85 &
48.98 & 53.95 & 58.46 & 59.95 &
18.60 & 23.92 & 25.74 & 28.55  \\ 
\midrule
\multirow{2}{*}{ALFA-Mix~\cite{parvaneh2022active}} & $G$ &
85.28 & 86.86 & 87.78 & 88.15 &
51.39 & 57.37 & 61.61 & 63.46 &
18.65 & 24.22 & \underline{26.32} & \underline{29.42} \\ 
& $L$ &
84.71 & 86.47 & 87.62 & 87.81 &
48.49 & 52.90 & 57.23 & 58.47 &
18.35 & 23.56 & 25.55 & 28.84  \\
\midrule
KAFAL~\cite{cao2023knowledge} & $G,L$ &
\underline{86.10} & \textbf{87.73} & \textbf{88.60} & \textbf{88.90} &
48.56 & 54.25 & 58.51 & 61.61 &
16.55 & 21.59 & 23.64 & 27.02 \\
\midrule
LoGo~\cite{kim2023re} & $G,L$ &
\textbf{86.15} & \underline{87.67} & 88.45 & 88.65 &
51.64 & 57.62 & 61.19 & 62.44 &
17.79 & 23.82 & 25.71 & 28.91 \\
\midrule
IFAL~\cite{ijcai2025p812} & $G,L$ &
82.94 & 84.57 & 85.82 & 86.09 &
45.60 & 50.78 & 55.55 & 57.67 &
17.72 & 22.77 & 24.37 & 27.81 \\
\midrule
\textbf{FairFAL (ours)} & $G,L$ &
85.81 & 87.48 & 88.43 & \textbf{88.90} &
\textbf{53.25} & \textbf{59.17} & \textbf{62.62} & \textbf{64.57} &
\textbf{18.89} & \textbf{24.45} & \textbf{26.47} & \textbf{30.02}  \\
\bottomrule
\end{tabular}
\caption{Test accuracy (\%) comparison on three benchmark datasets under different labeled-data ratios, with $\alpha = 100$ and $\rho = 20$. Results are averaged over five random seeds. Traditional AL baselines are evaluated using two query-selector models, $G$ and $L$, denoting global and local selectors, respectively. \textbf{Bold} numbers indicate the best result in each column, while \underline{underlined} numbers denote the second best.
}
\label{tab:main_numeric2}
\end{table*}

Corresponding to Figure~\ref{fig:main_results}, Tables~\ref{tab:main_numeric} and~\ref{tab:main_numeric2} provide the full numerical results across different labeled-data ratios. FairFAL consistently ranks among the top methods, and its advantage becomes increasingly evident as dataset complexity grows—from FMNIST to CIFAR-10 and further to CIFAR-100.
This trend stems from the varying reliance on global class balance. On simpler datasets, majority-class samples often exhibit high confidence and low uncertainty, allowing minority samples to be discovered even without explicit balancing. In contrast, more challenging datasets require stronger class-balance control: many majority-class samples remain uncertain, and querying them exacerbates global imbalance, whereas selecting minority samples with similar uncertainty yields notably greater benefit.

For uncertainty-driven strategies (Entropy, Margin, and Least Confidence), we observe that local-model querying performs better when $\alpha{=}0.1$, while global-model querying becomes preferable on CIFAR-10 and CIFAR-100 when $\alpha{=}100$. These patterns fully align with the analyses in Section~\ref{sec:observation}. Minor deviations appear on FMNIST under $\alpha{=}100$, likely because the lower task difficulty limits the benefit that the global model gains from cross-client aggregation.
For relatively simple tasks, a local-only querying strategy or a larger threshold $\delta$ in Eq.~\ref{eq:ms} may therefore be more suitable.
For diversity-based sampling (e.g., Coreset), the global model consistently achieves the best performance across all configurations, which again corroborates our earlier observations. Hybrid strategies such as BADGE, GCNAL, and ALFA-Mix combine uncertainty and diversity cues and do not exhibit a single dominant pattern.
Existing FAL methods (KAFAL, LoGo, and IFAL) lack mechanisms that explicitly promote class-balanced query selection. Consequently, they fail to achieve competitive performance and, in many cases, perform worse than standard AL strategies. These results further validate the necessity and effectiveness of the class-fair design introduced in FairFAL.


\begin{figure*}[!ht]
\centering
   \includegraphics[width=0.95\linewidth]{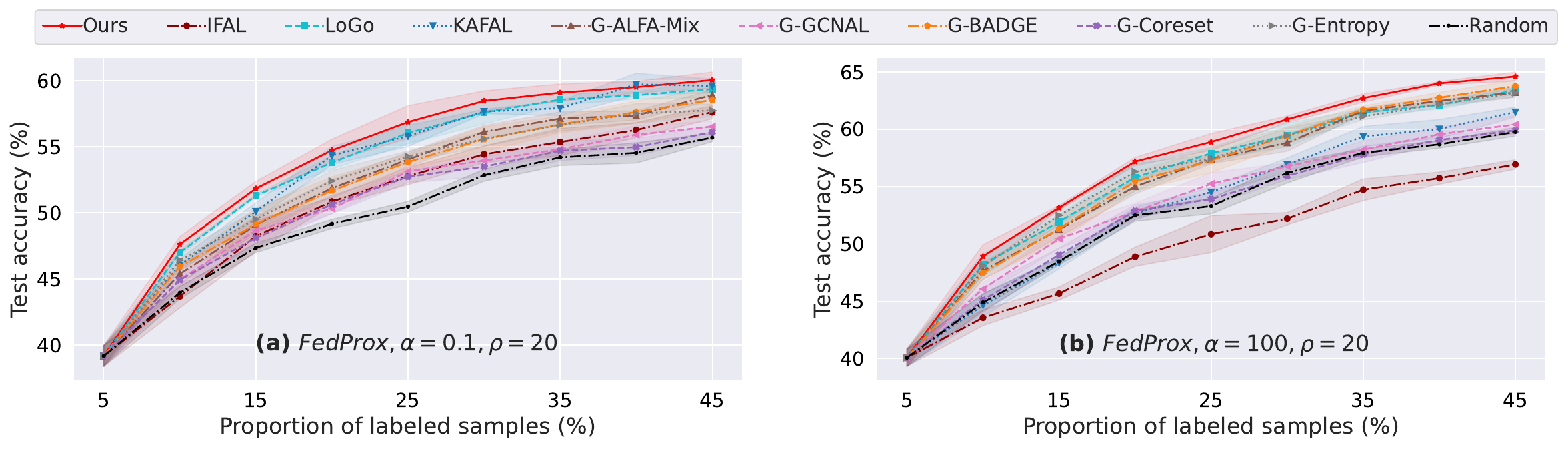}
   \includegraphics[width=0.95\linewidth]{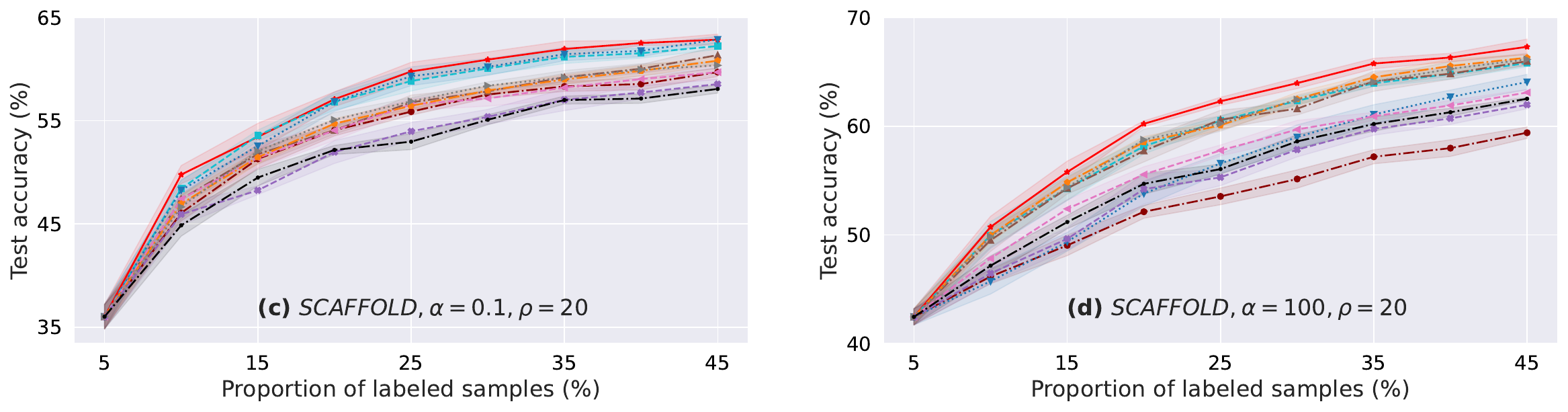}
   \includegraphics[width=0.95\linewidth]{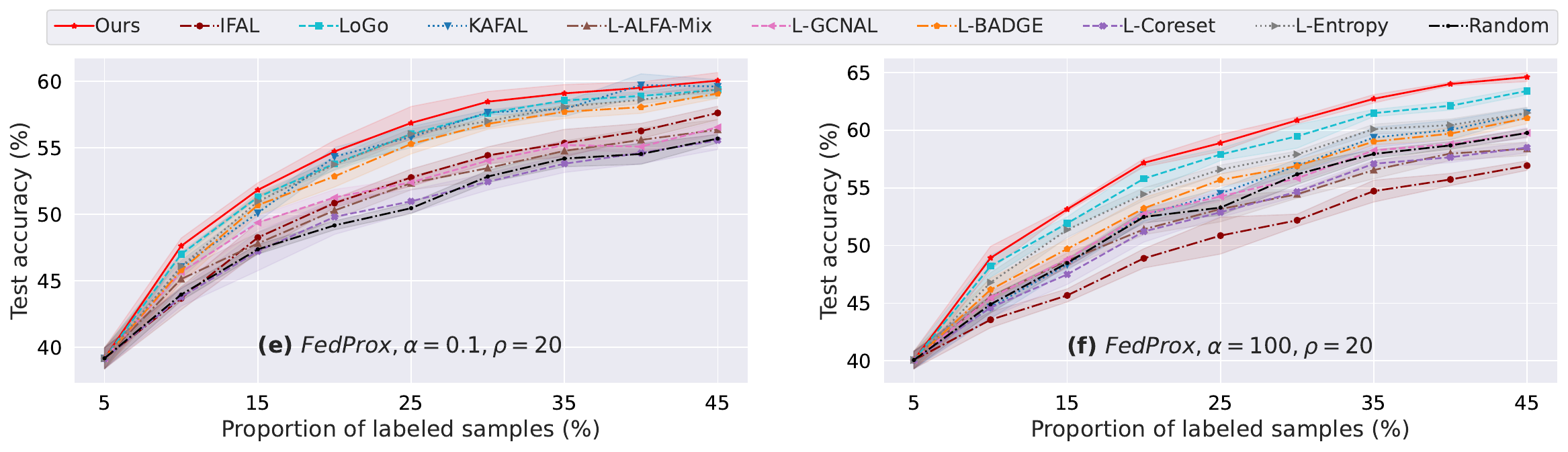}
   \includegraphics[width=0.95\linewidth]{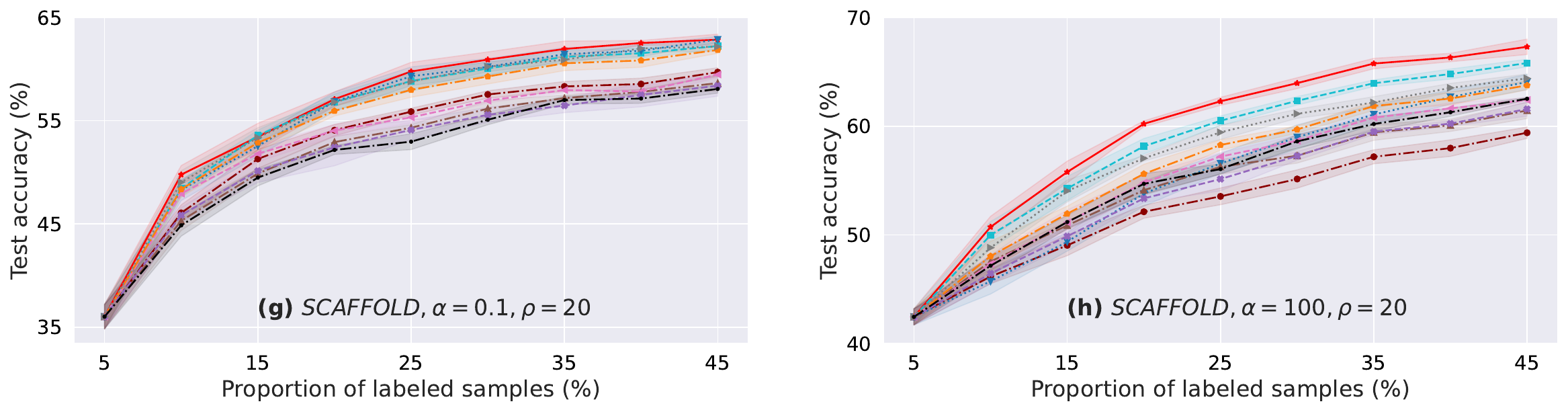}
\caption{Test accuracy versus labeled-data ratio on CIFAR-10 under global imbalance ($\rho{=}20$) and two heterogeneity levels ($\alpha{=}0.1$ and $\alpha{=}100$), evaluated with FedProx and SCAFFOLD.
Each subplot shows the corresponding FL framework and heterogeneity setting, comparing FairFAL with FAL and AL baselines using either the global (G-) or local (L-) query model.}
	\label{fig:gf_results}
\end{figure*}


\begin{figure*}[!ht]
\centering
   \includegraphics[width=0.95\linewidth]{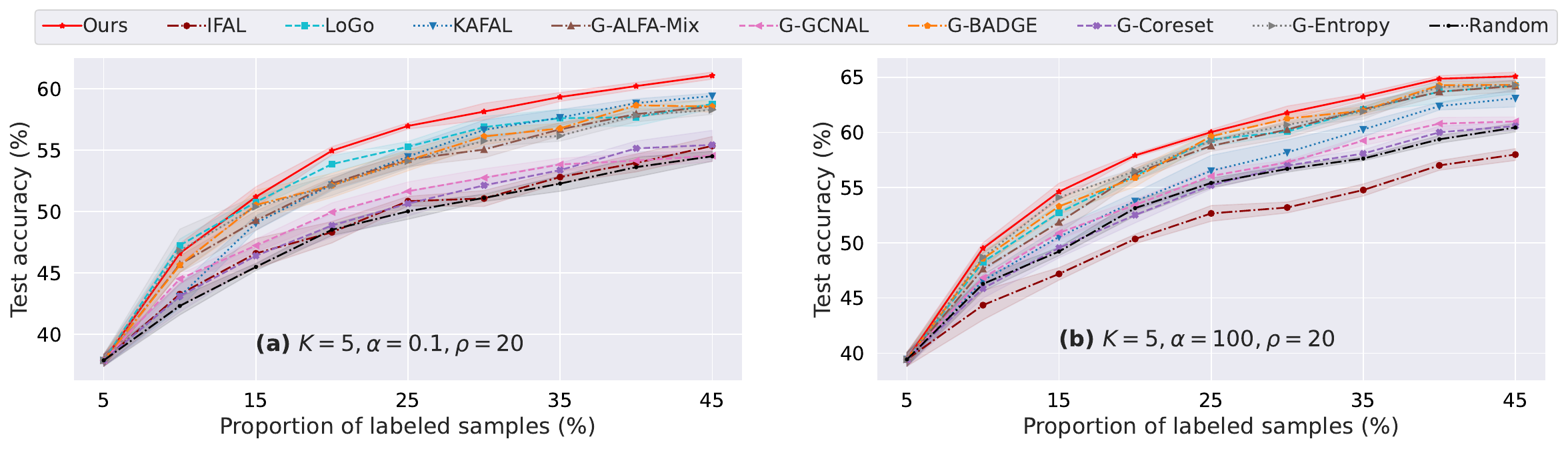}
   \includegraphics[width=0.95\linewidth]{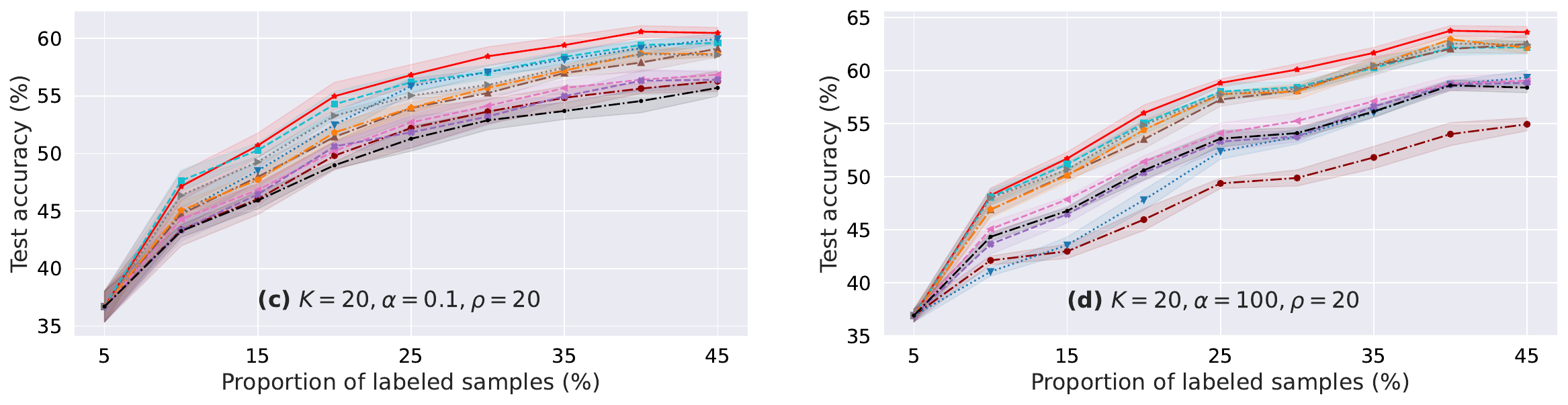}
   \includegraphics[width=0.95\linewidth]{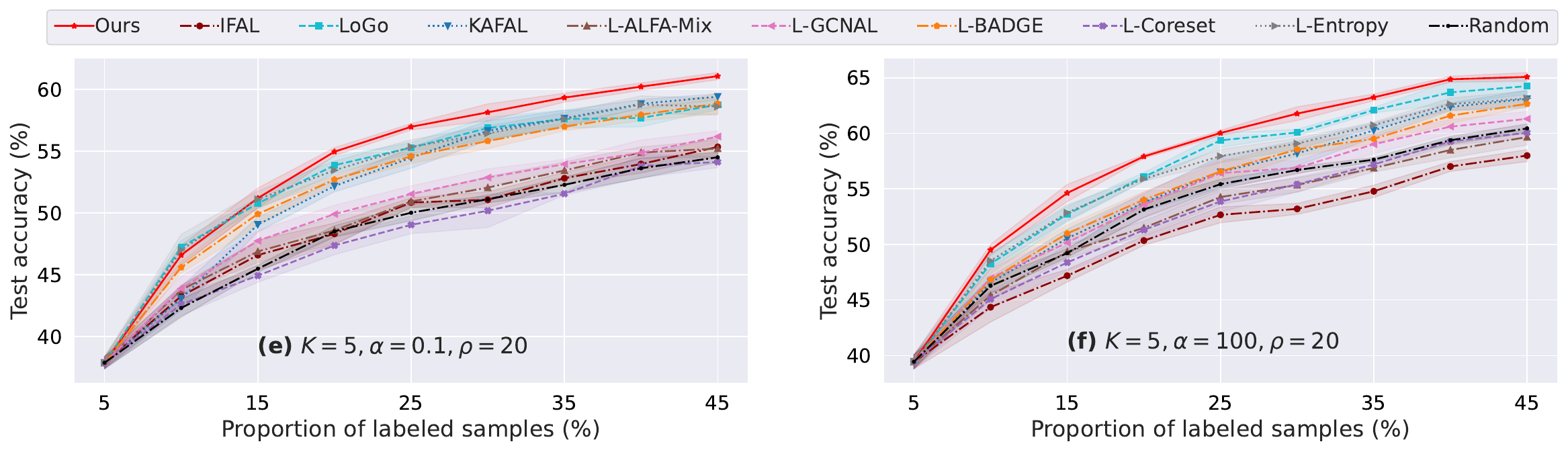}
   \includegraphics[width=0.95\linewidth]{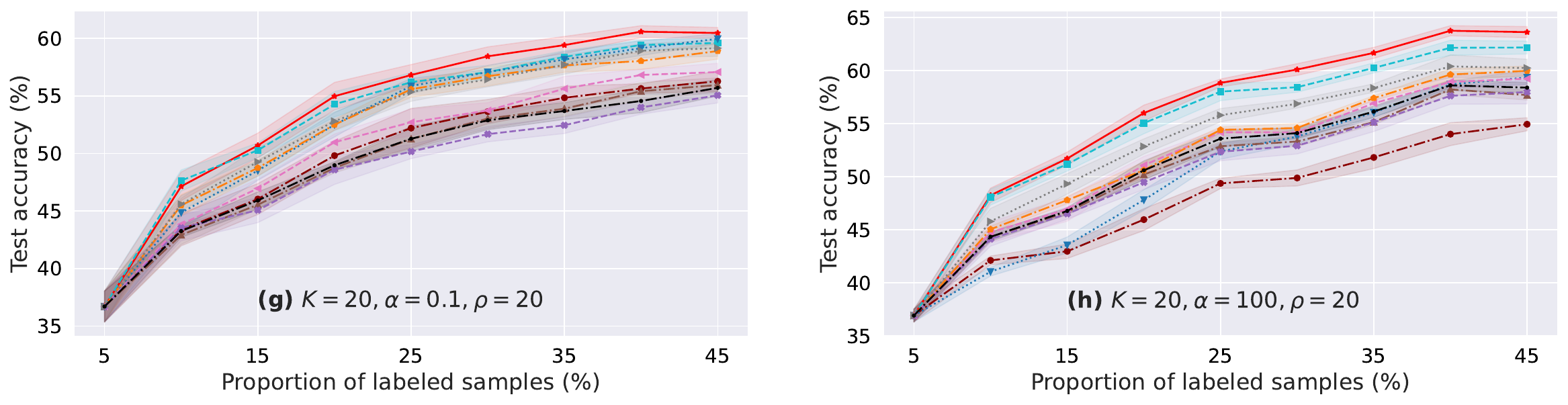}
\caption{Test accuracy versus labeled-data ratio on CIFAR-10 under global imbalance ($\rho{=}20$) and two heterogeneity levels ($\alpha{=}0.1$ and $\alpha{=}100$), evaluated with either $5$ or $20$ clients.
Each subplot corresponds to a specific combination of client count and heterogeneity level, and compares FairFAL with FAL and AL baselines using either the global (G-) or local (L-) query model.}
	\label{fig:gk_results}
\end{figure*}

\subsection{Additional Ablation Studies }

\textbf{Generalization across FL frameworks.}
We examine FairFAL under two widely used FL frameworks—FedProx~\cite{li2020federated} and SCAFFOLD~\cite{karimireddy2020scaffold}—to evaluate its robustness to different optimization dynamics. As illustrated in Figure~\ref{fig:gf_results}, FairFAL consistently surpasses all baselines across both frameworks and both heterogeneity settings. These results indicate that FairFAL operates orthogonally to the underlying FL optimizer and can be seamlessly incorporated into diverse FL pipelines. Such compatibility highlights its practical value for real-world deployments, where different FL frameworks may be selected based on system constraints or stability considerations.

\textbf{Scalability across client counts.}
We further assess the scalability of FairFAL by varying the number of participating clients to $5$ and $20$. As shown in Figure~\ref{fig:gk_results}, FairFAL consistently achieves strong performance in both cases, outperforming all baselines regardless of the extent of data fragmentation. These results indicate that FairFAL effectively preserves class balance and maintains high-quality query selection across different client counts, confirming its scalability and suitability for large-scale federated deployments.


\begin{table*}
\centering
\setlength{\tabcolsep}{8.5pt}
\renewcommand{\arraystretch}{1.15}
\begin{tabular}{l | cccc  cccc }
\toprule
\multirow{3}{*}{\textbf{Method}} & \multicolumn{8}{c}{\textbf{CIFAR-10}} \\
& \multicolumn{4}{c}{{$\alpha=0.1$, $\rho=20$}} &
\multicolumn{4}{c}{{$\alpha=100$, $\rho=20$}}  \\
\cmidrule(lr){2-5} \cmidrule(lr){6-9}
& 15\% & 25\% & 35\% & 45\% & 15\% & 25\% & 35\% & 45\% \\
\midrule
FairFAL (LC) &
51.64 & \underline{56.83} & \underline{59.22} & \textbf{60.56} &
\underline{53.17} & \textbf{59.65} & \textbf{63.25} & \underline{64.27}  \\
\midrule
FairFAL (Margin) &
\underline{51.78} & 56.82 & 58.94 & 60.48 &
52.82 & \underline{59.63} & \underline{63.05} & \textbf{64.57}  \\
\midrule
FairFAL (Entropy) &
\textbf{52.12} & \textbf{56.90} & \textbf{59.62} & 60.44 &
\textbf{53.25} & 59.17 & 62.62 & \textbf{64.57}   \\
\bottomrule
\end{tabular}
\caption{Test accuracy versus labeled-data ratio of FairFAL under different uncertainty measures on CIFAR-10 with $\alpha{=}0.1$ and $\alpha{=}100$ (global imbalance $\rho{=}20$). We replace the default entropy-based uncertainty with Margin and Least Confidence (LC) scoring. }
\label{tab:main_numeric3}
\end{table*}

\begin{table*}
\centering
\begin{tabular}{l | cccc  cccc }
\toprule
\multirow{3}{*}{\textbf{Method}} & \multicolumn{8}{c}{\textbf{CIFAR-10}} \\
& \multicolumn{4}{c}{{$\alpha=0.1$, $\rho=20$}} &
\multicolumn{4}{c}{{$\alpha=100$, $\rho=20$}}  \\
\cmidrule(lr){2-5} \cmidrule(lr){6-9}
& 15\% & 25\% & 35\% & 45\% & 15\% & 25\% & 35\% & 45\% \\
\midrule
FairFAL ($\delta=0.65$) &
\textbf{52.25} & \underline{56.89} & \underline{59.33} & \underline{60.27} &
\underline{53.25} & 59.17 & \underline{62.62} & \textbf{64.57}  \\
\midrule
FairFAL ($\delta=0.70$) &
\underline{52.12} & \textbf{56.90} & \textbf{59.62} & \textbf{60.44} &
\underline{53.25} & 59.17 & \underline{62.62} & \textbf{64.57}  \\
\midrule
FairFAL ($\delta=0.75$, \textbf{Final}) &
\underline{52.12} & \textbf{56.90} & \textbf{59.62} & \textbf{60.44} &
\underline{53.25} & 59.17 & \underline{62.62} & \textbf{64.57}   \\
\midrule
FairFAL ($\delta=0.80$) &
\underline{52.12} & \textbf{56.90} & \textbf{59.62} & \textbf{60.44} &
\textbf{53.47} & \underline{59.34} & 62.44 & 64.32   \\
\midrule
FairFAL ($\delta=0.85$) &
\underline{52.12} & \textbf{56.90} & \textbf{59.62} & \textbf{60.44} &
53.03 & \textbf{59.74} & \textbf{63.17} & \underline{64.55}   \\
\bottomrule
\end{tabular}
\caption{Test accuracy versus labeled-data ratio of FairFAL under different
model-selection thresholds $\delta$ on CIFAR-10 with $\alpha{=}0.1$
and $\alpha{=}100$ (global imbalance $\rho{=}20$). We evaluate
$\delta\!\in\!\{0.65,0.70,0.75,0.80,0.85\}$, where $\delta{=}0.75$
corresponds to the default configuration.}
\label{tab:main_numeric4}
\end{table*}

\textbf{Impact of different uncertainty measures.}
We further examine the robustness of FairFAL by replacing the default entropy-based uncertainty score with two widely used alternatives: Margin and Least Confidence (LC). 
As reported in Table~\ref{tab:main_numeric3}, all three variants achieve highly similar performance across different labeled data ratios and heterogeneity conditions, with entropy and least confidence showing a small but steady advantage in most cases. 
The performance differences remain minimal, typically within a range of approximately $0.3\%$ to $0.6\%$, demonstrating that FairFAL is not sensitive to the specific form of uncertainty scoring. 
This stability stems from its class-aware sampling and diversity-driven refinement. 
Once prototype-guided pseudo labeling establishes reliable class partitions, the uncertainty score mainly serves to rank samples within each class, allowing FairFAL to retain strong performance under a broad set of commonly adopted uncertainty measures.

\textbf{Impact of the threshold $\delta$.}
We also study the sensitivity of FairFAL to the model-selection threshold $\delta$, which determines whether a client relies on the global or local model for querying. As reported in Table~\ref{tab:main_numeric4}, sweeping $\delta$ from $0.65$ to $0.85$ produces only marginal performance changes. All settings yield nearly identical accuracy across labeled-data ratios and heterogeneity levels. The default choice $\delta{=}0.75$ achieves consistently strong results and remains competitive throughout, indicating that FairFAL does not depend on fine-grained parameter tuning. This stability reflects the reliability of the model-selection score itself, which already provides a clear distinction between conditions favoring global or local querying. As a result, FairFAL maintains effective class-balanced sampling even under substantial variations of $\delta$, underscoring its robustness and practical deployability in real federated environments.

\begin{table*}[t]
\centering
\begin{tabular}{l | cccc }
\toprule
\multirow{2}{*}{\textbf{Method}}
& \multicolumn{2}{c}{\textbf{MobileNet}} &
\multicolumn{2}{c}{\textbf{ResNet-18}}  \\
\cmidrule(lr){2-3} \cmidrule(lr){4-5}
& $\alpha=0.1$ & $\alpha=100$ & $\alpha=0.1$ & $\alpha=100$ \\
\midrule
KAFAL &
\textbf{41.16} & 43.14 & \underline{37.96} & 42.57  \\
LoGo &
39.08 & \underline{45.87} & 37.10 & \underline{44.95}  \\
IFAL &
36.64 & 41.80 & 36.89 & 42.10   \\
\midrule
\textbf{FairFAL} &
\underline{41.07} & \textbf{46.66} & \textbf{38.42} & \textbf{45.53}   \\
\bottomrule
\end{tabular}
\caption{Final-round test accuracy (seed = 1) of KAFAL, LoGo, IFAL, and FairFAL under different backbone architectures on CIFAR-10 with heterogeneity levels $\alpha=0.1$ and $\alpha=100$ (global imbalance $\rho=20$). The default 4-layer CNN backbone is replaced with MobileNet and ResNet-18, respectively.}
\label{tab:main_numeric5}
\end{table*}

\textbf{Impact of backbone architecture.}
To evaluate the architectural robustness of FairFAL, we replace the default 4-layer CNN with two stronger backbones, MobileNet~\cite{howard2017mobilenets} and ResNet-18~\cite{he2016deep}, and compare its performance against existing FAL methods. As presented in Table~\ref{tab:main_numeric5}, FairFAL consistently matches or outperforms existing FAL methods across both architectures and both heterogeneity levels. In contrast, methods such as KAFAL, LoGo, and IFAL exhibit notable fluctuations when the backbone changes. FairFAL’s stable performance demonstrates that its core mechanisms do not depend on any particular network design and generalize effectively to modern, deeper architectures commonly used in practical federated systems.


\end{document}